\documentclass[letterpaper]{article} 
\usepackage[preprint]{aaai2027}
\usepackage[hyphens]{url}  
\usepackage{graphicx} 
\usepackage{natbib}  
\usepackage{caption} 
\usepackage{amssymb}
\usepackage{amsmath}
\usepackage{booktabs}
\usepackage{pifont}
\newcommand{\cmark}{\ding{51}}
\newcommand{\xmark}{\ding{55}}
\usepackage{array}
\usepackage{multirow}
\usepackage{makecell}
\providecommand{\corresponding}{\thanks{Corresponding author.}}

\title{SAGE: Subpopulation-Aware Generative Enhancement for \\Mitigating Spurious Correlations}
\author{
    Yiming Luo\textsuperscript{\rm 1},
    Rongqiang Zhao\textsuperscript{\rm 1,\rm 2}\corresponding,
    Jie Liu\textsuperscript{\rm 1,\rm 2}
}
\affiliations{
    \textsuperscript{\rm 1}Harbin Institute of Technology\\
    \textsuperscript{\rm 2}State Key Laboratory of Smart Farm Technologies and Systems\\
    2025112368@stu.hit.edu.cn, zhaorq@hit.edu.cn
}

\begin{document}

\maketitle

\begin{abstract}
Spurious correlations pose a significant challenge to the robustness of modern machine learning. The inherent imbalance in dataset distributions often leads traditional Empirical Risk Minimization (ERM) models to rely on majority spurious attributes for classification, resulting in poor performance on minority groups. This problem becomes particularly challenging when the spurious attributes are unavailable. Existing group-label-free methods often upsample minority groups or misclassified real training examples; repeating the same instances can reduce effective diversity and encourage overfitting. To mitigate these spurious correlations from a data-centric perspective in the absence of prior knowledge, we introduce Subpopulation-Aware Generative Enhancement (SAGE), a two-stage generative augmentation framework. Using cluster-derived sub-labels and class labels, we adapt a conditional diffusion model via LoRA alongside learnable sub-label token embeddings, generating targeted synthetic data to fill underrepresented regions in the training set and construct a sub-label-balanced synthetic reweighting set for last-layer reweighting. We experimentally show that SAGE achieves 89.5\%, 85.7\%, and 79.1\% worst-group accuracy on Waterbirds, CelebA, and MetaShift, respectively, outperforming the best group-label-free baselines by up to 7.7 percentage points.
\end{abstract}
\begin{links}
    \link{Code}{https://github.com/luoym-lym/SAGE}
\end{links}

\section{Introduction}
\label{Intro}
\begin{figure*}[t]
    \centering
    \includegraphics[width=\textwidth]{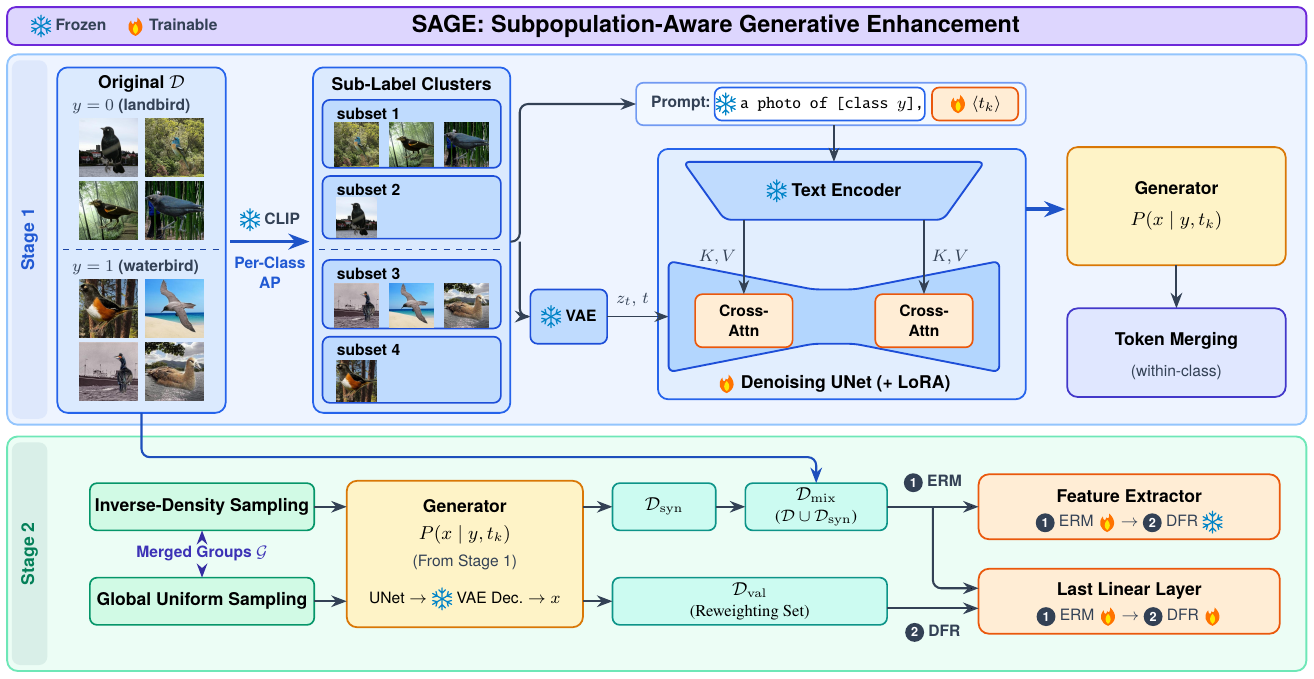}
    \caption{Overview of SAGE. \textbf{Stage 1:} Within-class CLIP+AP clustering assigns sub-label tokens $\langle t_k \rangle$ to adapt Stable Diffusion (UNet LoRA and learnable token embeddings) into $P(x \mid y, t_k)$, followed by within-class token merging. \textbf{Stage 2:} Inverse-density sampling synthesizes $\mathcal{D}_{\mathrm{syn}}$ for ERM on $\mathcal{D}_{\mathrm{mix}} = \mathcal{D} \cup \mathcal{D}_{\mathrm{syn}}$ (step~1), followed by uniform sampling of $\mathcal{D}_{\mathrm{val}}$ for last-layer DFR with a frozen feature extractor (step~2).}
    \label{fig:framework}
\end{figure*}
Deep neural networks have achieved outstanding performance in classification tasks \cite{krizhevsky2012imagenet, he2016resnet}. However, they often tend to rely on easily accessible shortcut features rather than the truly causal core features \cite{geirhos2020shortcut}. For example, in the task of classifying waterbirds and landbirds, if the majority of waterbirds in the training set appear against water backgrounds while most landbirds appear against land backgrounds, the model will inherently adopt the background (water or land) as its primary basis for classification. Consequently, when such spurious correlations are broken during the testing phase (e.g., encountering a waterbird on land), the model suffers a catastrophic drop in test performance, which is primarily manifested as low worst-group accuracy (WGA) \cite{beery2018recognition,sagawa2020distributionally}.

To mitigate spurious correlations, existing methods can be broadly divided into two categories. The first category assumes access to spurious group information and improves robustness by modifying the model's learning objective or reweighting strategy \cite{sagawa2020distributionally,kirichenko2023last}. The second category avoids group labels and reduces model bias using proxy signals inferred from model behavior, input regions, or discovered concepts \cite{liu2021just,nam2020learning,zhang2022correct,noohdani2024decompose,asgari2022masktune,wu2023discover}.

Recently, generative models have been increasingly applied to data augmentation across various domains \cite{azizi2023synthetic,dunlap2023diversify,sun2026ose}, and a nascent line of work has begun to explore their potential for mitigating spurious correlations. This line of work uses image generation to create additional samples under different visual conditions, class-attribute combinations, or debiased prompts \cite{qraitem2024ffr,ghosh2024aspire,basu2026harnessing,parast2025ddb}. Compared with reusing existing training examples, generated data can provide new visual instances and thus offers a natural way to expand the available training set.

Taken together, these directions leave a gap when spurious attributes are unobserved and manually curated group-balanced validation sets are unavailable, a stricter setting that more closely matches real-world scenarios than assuming access to group annotations or curated validation data. Existing group-label-free approaches still lag in worst-group accuracy, whereas prior generative methods often depend on external models (e.g., LLMs or segmentation tools) to discover spurious attributes, explicit group labels to guide generation, or manually curated validation sets for model selection.

To overcome these limitations, we propose SAGE (Figure~\ref{fig:framework}), a data-centric generative augmentation framework requiring neither external attribute-discovery tools (e.g., LLMs or segmentation models), manually annotated spurious groups, nor curated validation sets. Our key insight is that visually salient spurious cues often structure the feature space of pretrained vision encoders: samples sharing similar shortcut features tend to cluster together even when spurious attributes are unlabeled. We therefore treat cluster assignments as proxies for bias-relevant subpopulations, rather than as ground-truth spurious groups.

SAGE instantiates this idea in two stages. First, we cluster within-class CLIP features, assign each cluster a learnable sub-label token to adapt a conditional generative model alongside the class label, and merge near-redundant tokens. Second, we use inverse-density sampling over merged groups to generate targeted training data for underrepresented regions, and uniform sampling to construct a sub-label-balanced synthetic reweighting set. Serving as a practical proxy for the balanced data required by Deep Feature Reweighting (DFR), this synthetic set enables last-layer retraining without manual curation.

We evaluate SAGE on Waterbirds, CelebA, and MetaShift, achieving 89.5\%, 85.7\%, and 79.1\% WGA, respectively, outperforming all prior methods without group labels or curated validation data while remaining competitive with group-aware methods. Additional ISIC medical imaging results and extended ablations appear in the Technical Supplement (Appendix~E). Our ablation and visualization studies show complementary benefits from training augmentation and synthetic reweighting, and confirm that learned sub-label tokens capture subpopulation-specific visual factors.

\textbf{Summary of contributions.} We summarize our contributions as follows:
\begin{itemize}
\item We propose SAGE, a generative augmentation method for mitigating spurious correlations in the practical setting where spurious attributes are unobserved and group-balanced validation data are unavailable.

\item We design a subpopulation-aware data construction strategy that derives sub-labels through within-class semantic feature clustering and couples them with complementary label sampling strategies, enabling targeted generation for underrepresented training regions and a sub-label-balanced synthetic reweighting set.

\item We provide ablation, visualization, and extended analyses showing the complementary roles of training-set augmentation and synthetic reweighting, as well as the ability of learned sub-label tokens to capture subpopulation-specific visual factors that often align with spurious attributes.
\end{itemize}

\section{Related Work}
\label{sec:related_work}
\textbf{Robust Learning for Spurious Correlations.}
A large body of work mitigates spurious correlations by modifying the training or model-selection procedure. Some methods use spurious group information directly: gDRO \cite{sagawa2020distributionally} dynamically upweights high-loss groups during training, and DFR \cite{kirichenko2023last} retrains the last linear layer on a group-balanced validation set. Another line of work avoids training-time group labels and instead relies on proxy signals. JTT \cite{liu2021just} and LfF \cite{nam2020learning} identify hard or biased examples from model behavior; CnC \cite{zhang2022correct} uses contrastive learning to improve robustness; and DaC \cite{noohdani2024decompose} addresses spurious correlations through a decompositional training strategy. More restrictive group-label-free methods include MaskTune \cite{asgari2022masktune}, which masks the most discriminative regions found by a trained model and fine-tunes on the masked images to force exploration of alternative cues, and DISC \cite{wu2023discover}, which discovers unstable human-interpretable concepts across environments and intervenes on the training data to reduce their influence.

\textbf{Generative Data Augmentation for Spurious Correlations.}
Diffusion models have recently become a powerful tool for controllable image generation \cite{ho2020denoising,dhariwal2021diffusion,rombach2022high}, motivating their use as synthetic data sources for robust classification \cite{azizi2023synthetic,dunlap2023diversify}. For mitigating spurious correlations, existing work has mainly explored how to construct useful synthetic samples for downstream classifiers. One line of work balances synthetic data before real-data training; for example, FFR \cite{qraitem2024ffr} pretrains on balanced synthetic images generated with prompts naming known spurious attributes and evaluates on CelebA-HQ (smiling vs.\ gender) under varied bias ratios rather than Waterbirds or standard CelebA (blond hair), precluding a direct row in Table~\ref{tab:generative_comparison}. Another line introduces additional guidance to make generation more targeted: ASPIRE \cite{ghosh2024aspire} uses LLMs and language-guided image editing to identify spurious features from captions, then personalizes a text-to-image model to generate non-spurious images; DDB \cite{parast2025ddb} combines textual inversion, language-based segmentation, diffusion generation, and ERM-based pruning to synthesize samples that break spurious correlations. Closely related to subpopulation-level generation, Clustered DreamBooth \cite{basu2026harnessing} clusters samples within known groups and fine-tunes DreamBooth models for each cluster to generate group-balanced data. Overall, these methods show the promise of generated data, but deciding what to generate often involves additional machinery, such as language-based attribute discovery, segmentation-assisted localization, concept banks, group-aware sampling rules, or validation-set-based model selection. In contrast, SAGE requires only a general-purpose visual encoder to turn within-class clusters into learnable tokens capturing unnamed shared visual factors, and reuses a single adapted generator for both training data and a synthetic reweighting set.

\section{Method}
In this section, we detail our proposed framework following the two-stage sequence: Stage~1 performs within-class clustering and conditional generator adaptation (followed by within-class token merging), while Stage~2 encompasses targeted data generation and downstream model training.
\subsection{Stage 1: Semantic Clustering and Conditional Generator Fine-Tuning}

\textbf{AP Clustering.} In the first stage of our framework, independently within each class $y$, we partition the training data into feature-space subgroups without access to explicit spurious attribute labels. We first map the training images into an L2-normalized latent feature space $\mathcal{Z} = \{z_1, \dots, z_N\}$ using a frozen CLIP ViT-B/32 image encoder \cite{radford2021learning}, and subsequently cluster them using Affinity Propagation (AP) \cite{frey2007clustering}. Unlike standard methods such as K-Means, AP does not require a pre-defined number of clusters; instead, it identifies representative samples (\textit{exemplars}) through an iterative message-passing mechanism. The algorithm operates on a similarity matrix defined by the negative squared Euclidean distance, $s(i, k) = -\|z_i - z_k\|^2$, where the diagonal preferences $s(k, k)$ (set to the mean off-diagonal similarity) and damping factor $\lambda_{\mathrm{damp}}$ control the resulting cluster granularity (Appendix~C). AP iteratively exchanges two types of damped messages between data points: \textit{responsibility} $r(i, k)$, which quantifies the suitability of $z_k$ serving as an exemplar for $z_i$, and \textit{availability} $a(i, k)$, which reflects the accumulated evidence for $z_i$ choosing $z_k$. Upon convergence, each data point $z_i$ is assigned to \emph{exactly one} exemplar $k$ maximizing $r(i, k) + a(i, k)$. On large datasets like CelebA ($N \approx 163\text{k}$), AP runs per class on a $K = 10{,}000$ Farthest Point Sampling core set and assigns all samples to their nearest exemplar via an exact Faiss index, reducing dominant memory from $O(N^2)$ to $O(ND + K^2)$ (Appendix~C); Waterbirds and MetaShift run exact per-class AP. We use these assignments as discrete sub-labels for generative conditioning, treating clusters as proxies for bias-relevant subpopulations rather than ground-truth spurious groups; post-hoc cluster purity reaches 93.7\%, 98.2\%, and 72.5\% on Waterbirds, CelebA, and MetaShift (Appendix~C).

\textbf{Fine-Tuning the Diffusion Model with LoRA.} We adopt Stable Diffusion v1.5 (SD) \cite{rombach2022high,ho2020denoising,dhariwal2021diffusion} as our conditional text-to-image backbone owing to its synthesis quality and open-source availability. To inject the discovered latent sub-labels into generation via cross-attention without the prohibitive cost of full fine-tuning, we adapt the UNet using Low-Rank Adaptation (LoRA) \cite{hu2022lora} and register a unique semantic token $\langle t_k \rangle$ for each sub-label $k$ in the text encoder's vocabulary, initialized from the word \textit{`photo'} as a domain-general prior.

During adaptation, we use the structured prompt template:
\begin{center}
    \textit{``a photo of [class label $y$], [sub-label token $\langle t_k \rangle$]''}
\end{center}
where [class label $y$] anchors the class semantics and $\langle t_k \rangle$ captures fine-grained subpopulation variation. These newly added token embeddings are jointly optimized with the UNet LoRA parameters, while the remaining text-encoder, VAE, and base UNet weights stay frozen.

Concretely, let $(x, y, k) \sim \mathcal{D}$ be a training image belonging to class $y$ and assigned to the latent cluster $k$. We map $x$ into the latent space using the frozen VAE encoder, denoted as $z = \mathcal{E}(x)$. The structured prompt is encoded by $\tau$---where $y$ is inserted as fixed textual class names and only the embedding $e_k$ of $\langle t_k \rangle$ is learnable---to yield the conditioning vector $c = \tau(y, e_k)$. By introducing trainable LoRA parameters $\Delta\theta$ into the frozen UNet weights $\theta$, we formulate the joint optimization objective as the following latent diffusion loss:
\begin{equation}
\begin{aligned}
    \mathcal{L}_{\mathrm{joint}}(\Delta\theta, \{e_k\}) &= \mathbb{E}_{(x,y,k)\sim\mathcal{D},\,\epsilon \sim \mathcal{N}(0,\mathbf{I}),\,t} \\
    &\quad \left[ \big\| \epsilon - \epsilon_{\theta+\Delta\theta}\big(z_t, t, \tau(y, e_k)\big) \big\|_2^2 \right]
\end{aligned}
\label{eq:joint_loss}
\end{equation}
where $t \sim \mathcal{U}(1, T)$ is the timestep, $\epsilon$ is the unscaled Gaussian noise, and $z_t$ is the noisy latent at step $t$. By minimizing $\mathcal{L}_{\mathrm{joint}}$, the network simultaneously learns to manipulate generative trajectories via $\Delta\theta$ and optimize $e_k$; during sampling, the UNet iteratively denoises a Gaussian latent conditioned via cross-attention (prompt embeddings as keys/values, image features as queries) before VAE decoding (Appendix~B).

\subsection{Stage 2: Targeted Data Generation and Downstream Model Training}
\textbf{Two Label Sampling Strategies for Data Generation.} With the joint conditional generator optimized in the previous stage, we can synthesize auxiliary samples to augment the original dataset. However, selecting appropriate condition combinations remains a critical challenge to effectively mitigate dataset biases. We employ an inverse-density sampling strategy for the condition labels. Let $N_k$ denote the number of training samples assigned to cluster $k$ and $q$ the number of clusters in that class. If inverse-density weights were normalized only within each class (before token merging), the per-class weight for sub-label $\langle t_k \rangle$ would be
\begin{equation}
    P_{\mathrm{train}}(\langle t_k \rangle) = \frac{N_k^{-1}}{\sum_{j=1}^{q} N_j^{-1}}.
\end{equation}
However, although AP assigns each sample to exactly one cluster, visually similar subpopulations within a class can still split across multiple fine-grained clusters, resulting in an inflated total sampling probability for redundant concepts under naive inverse-density sampling. Therefore, before inverse-density sampling, we merge nearly redundant sub-label tokens within each class. After Stage~1 fine-tuning, we compute pairwise cosine similarities between the learned text-encoder embeddings $\{e_k\}$ in the same class; token pairs exceeding $\beta \cdot \bar{s}_y$ ($\bar{s}_y$ is the within-class mean similarity; $\beta = 2.5, 4.0, 4.0$ on Waterbirds, CelebA, and MetaShift, fixed without group labels or test feedback; Appendix~B) are merged transitively into disjoint groups within each class. Let $\mathcal{G}$ denote the collection of all merged groups across classes. Because AP clustering and token merging are performed independently within each class label $y$ (Appendix~C), each merged group $G_m \in \mathcal{G}$ still belongs to a single class (no cross-class sharing or merging), and only the denominator normalizes training-set sampling probabilities globally over all sub-labels:
\begin{equation}
    \label{sample_strategy}
    P_{\mathrm{train}}(\langle t_k \rangle) = \frac{1}{|G_m|} \cdot \frac{N(G_m)^{-1}}{\sum_{G' \in \mathcal{G}} N(G')^{-1}}
\end{equation}
where $N(G_m) = \sum_{i \in G_m} N_i$ denotes the total empirical sample size of the merged group $G_m$ in the original training set, $|G_m|$ represents the number of distinct sub-labels contained within $G_m$, and the sum runs over all merged groups in $\mathcal{G}$. Because sub-labels only proxy bias-relevant subpopulations, inverse-density sampling upweights empirically rare clusters rather than verified worst-case class-attribute groups; some low-density clusters may reflect non-spurious variation (e.g., pose or lighting). Crucially, SAGE never discards clusters and retains all real data ($\mathcal{D}_{\mathrm{mix}} = \mathcal{D} \cup \mathcal{D}_{\mathrm{syn}}$ with $Q = |\mathcal{D}|$ generated images), so augmenting rare non-spurious appearances can still encourage class features invariant across them. We use global normalization in part to account for potential long-tailed structure; on CelebA, for example, the not-blond class is much larger than the blond class in the standard benchmark split \cite{sagawa2020distributionally}.

Furthermore, to fully unlock the framework's potential without manual data curation, we reuse the optimized generator to construct a balanced \emph{synthetic reweighting set} $\mathcal{D}_{\mathrm{val}}$ for Deep Feature Reweighting (DFR) \cite{kirichenko2023last}. Our goal is globally uniform coverage of merged groups: after the same within-class token merging as in training, we assign each merged group the same number of synthetic images (`samples-per-token'), rather than stochastic sampling over clusters or groups. This yields a deterministic, globally balanced reweighting set; we clearly distinguish the original training set $\mathcal{D}$, held-out benchmark validation splits (unused by SAGE), and $\mathcal{D}_{\mathrm{val}}$ (used solely to fit the final linear layer). Table~\ref{tab:main_results} uses `samples-per-token' $= 16 / 16 / 6$ on Waterbirds, CelebA, and MetaShift, fixed a priori before ablation without group labels or test feedback (Appendix~B).

\textbf{Training the Model and Deep Feature Reweighting.}
We utilize the newly constructed mixed training set $\mathcal{D}_{\mathrm{mix}}$ to train the downstream classifier, which consists of a feature extractor followed by a last linear layer. To strictly isolate the performance gains attributed to our augmentation framework, we adhere to the standard Empirical Risk Minimization (ERM) paradigm without introducing any modifications to the network architecture, objective function, or data sampling heuristics.

After the initial ERM training on $\mathcal{D}_{\mathrm{mix}}$, we apply Deep Feature Reweighting (DFR) \cite{kirichenko2023last}. DFR is based on the observation that an ERM-trained feature extractor can still encode core features, even when the final classifier relies heavily on spurious cues. In the original DFR setting, the feature extractor is frozen and the last classification layer is retrained from scratch on a small reweighting set balanced across spurious groups, typically constructed from group-labeled validation data. In our setting, true spurious groups are unavailable, so we use the sub-label-balanced synthetic reweighting set $\mathcal{D}_{\mathrm{val}}$ as a proxy. Retraining only the last linear layer on $\mathcal{D}_{\mathrm{val}}$ allows SAGE to approximate the group-balanced DFR procedure without manually curated class-attribute-balanced data. Further implementation details are elaborated in Appendix~B.

\begin{table*}[t]
\centering
\caption{Comparison of worst-group accuracy and average accuracy across three datasets. SAGE and Base (ERM) report the mean and standard deviation across three independent runs; literature baselines follow their source reports (JTT and LfF entries without $\pm$ omit standard deviations in their original publications). The Group Info column indicates whether group labels are used, where \cmark\cmark denotes that validation group labels are also utilized during training. Bold and underlined numbers indicate the best results within each group and globally, respectively.}
\label{tab:main_results}
\begin{tabular}{llccccccc}
\toprule
Category & Method & Group Info & \multicolumn{2}{c}{Waterbirds} & \multicolumn{2}{c}{CelebA} & \multicolumn{2}{c}{MetaShift}\\
\cmidrule(lr){4-5} \cmidrule(lr){6-7} \cmidrule(lr){8-9}
& & train/val & Worst & Average & Worst & Average & Worst & Average\\
\midrule
\multirow{2}{*}{\makecell[l]{Labels in\\train \& val}} & DFR & \xmark~/\cmark\cmark & $\underline{\mathbf{92.9_{\pm0.2}}}$ & $\mathbf{93.3_{\pm0.5}}$ & $88.3_{\pm1.1}$ & $91.3_{\pm0.3}$ & $\mathbf{72.8_{\pm0.6}}$ & $\mathbf{77.5_{\pm0.6}}$ \\
& gDRO & \cmark~/\cmark & $89.9_{\pm0.6}$ & $92.0_{\pm0.6}$ & $\underline{\mathbf{88.9}_{\pm1.3}}$ & $\mathbf{93.9_{\pm0.1}}$ & - & - \\
\midrule
\multirow{4}{*}{\makecell[l]{Labels in val}} & JTT & \xmark~/\cmark & 86.7 & 93.3 & 81.1 & 88.0 & $64.6_{\pm2.3}$ & $74.4_{\pm0.6}$ \\
& LfF & \xmark~/\cmark & 78.0 & 91.2 & 77.2 & 85.1 & - & - \\
& CnC & \xmark~/\cmark & $88.5_{\pm0.3}$ & $90.9_{\pm0.1}$ & $\mathbf{88.8_{\pm0.9}}$ & $89.9_{\pm0.5}$ & - & - \\
& DaC & \xmark~/\cmark & $\mathbf{92.3_{\pm0.4}}$ & $\underline{\mathbf{95.3}_{\pm0.4}}$ & $81.9_{\pm0.7}$ & $\mathbf{91.4_{\pm1.1}}$ & $\mathbf{78.3_{\pm1.6}}$ & $\mathbf{79.3_{\pm0.1}}$ \\
\midrule
\multirow{6}{*}{\makecell[l]{No group labels}} & JTT & \xmark~/\xmark & - & - & $40.6$ & $88.0$ & - & - \\
& LfF & \xmark~/\xmark & - & - & $24.4$ & $85.1$ & - & - \\
& MaskTune & \xmark~/\xmark & $86.4_{\pm1.9}$ & $93.0_{\pm0.7}$ & $78.0_{\pm1.2}$ & $91.3_{\pm0.1}$ & $66.3_{\pm6.3}$ & $73.1_{\pm2.2}$ \\
& DISC & \xmark~/\xmark & $88.7_{\pm0.4}$ & $\mathbf{93.8_{\pm0.7}}$ & - & - & $73.5_{\pm1.4}$ & $75.5_{\pm1.1}$ \\
& SAGE (ours) & \xmark~/\xmark & $\mathbf{89.5}_{\pm0.1}$ & $93.5_{\pm0.1}$ & $\mathbf{85.7_{\pm0.3}}$ & $90.1_{\pm0.2}$ & $\underline{\mathbf{79.1_{\pm1.8}}}$ & $\underline{\mathbf{82.3_{\pm0.2}}}$ \\
& Base (ERM) & \xmark~/\xmark & $73.2_{\pm4.2}$ & $90.8_{\pm0.5}$ & $44.6_{\pm0.3}$ & $\underline{\mathbf{95.5_{\pm0.2}}}$ & ${64.2_{\pm3.3}}$ & $77.4_{\pm1.8}$ \\
\bottomrule
\end{tabular}
\end{table*}
\section{Experiment}
In this section, we present our detailed experimental setup and results, and conduct ablation studies.
\label{experiment}
\subsection{Experiment setup}
\textbf{Datasets}. We evaluate our method on three standard benchmarks spanning ~1K to ~163K training images (Appendix~A): Waterbirds \cite{welinder2010caltech,sagawa2020distributionally}, CelebA \cite{liu2015deep} and MetaShift \cite{liang2022metashift}. Each dataset pairs a class label $y_i \in \mathcal{Y}$ with a spurious attribute $a_i \in \mathcal{A}$ to form $(y, a)$ groups; we evaluate average accuracy and worst-group accuracy (WGA), defined as the lowest test accuracy across all $(y, a)$ groups.

\textit{Waterbirds} \cite{welinder2010caltech,sagawa2020distributionally}: For the Waterbirds dataset, $\mathcal{Y}=\{\text{waterbird}, \text{landbird}\}$ and $\mathcal{A}=\{\text{water background}, \text{land background}\}$. 95\% of the training images have the same bird type $\mathcal{Y}$ and background type $\mathcal{A}$.

\textit{CelebA} \cite{liu2015deep}: In this dataset, $\mathcal{Y}=\{\text{blond hair}, \text{not blond hair}\}$ and $\mathcal{A}=\{\text{male}, \text{female}\}$. Only 6\% of blond celebrities in the dataset are male.

\textit{MetaShift} \cite{liang2022metashift}: Our setup for this dataset follows \cite{wu2023discover}, $\mathcal{Y}=\{\text{dog}, \text{cat}\}$ and $\mathcal{A}=\{\text{sofa}, \text{bed}, \text{bench}, \text{bike}\}$. In the training set, ``cat'' co-occurs with ``sofa'' and ``bed'', while ``dog'' co-occurs with ``bench'' and ``bike''; test images appear exclusively on unseen ``shelf'' backgrounds (Appendix~A). An additional evaluation on ISIC medical imaging is provided in Appendix~E.

\textbf{Basic Models}. For a fair comparison, similar to other compared methods, we use a ResNet-50 model \cite{he2016resnet} pre-trained on ImageNet as the classifier, and training hyperparameters are the same as \cite{sagawa2020distributionally}. In our method, we use CLIP ViT-B/32 \cite{radford2021learning} as the semantic embedding model for clustering. Furthermore, we employ the Hugging Face implementation of Stable Diffusion v1.5\footnote{\url{https://huggingface.co/stable-diffusion-v1-5/stable-diffusion-v1-5}} as the foundational generative model.

\textbf{Compared Methods}. We compare against non-generative approaches grouped by prior-knowledge usage: (1) explicit group labels during training (gDRO \cite{sagawa2020distributionally}) or last-layer reweighting on a group-balanced validation set (DFR \cite{kirichenko2023last}); (2) validation-only group labels for model selection (JTT \cite{liu2021just}, LfF \cite{nam2020learning}, CnC \cite{zhang2022correct}, DaC \cite{noohdani2024decompose}); and (3) strictly no group labels (MaskTune \cite{asgari2022masktune}, DISC \cite{wu2023discover}, and SAGE). We also compare against generative augmentation baselines: Clustered DreamBooth \cite{basu2026harnessing}, ASPIRE \cite{ghosh2024aspire}, and DDB \cite{parast2025ddb} (see Related Work). All comparisons follow standard splits, ImageNet-pretrained ResNet-50, and established protocols; we verified protocol consistency for open-source baselines and quote published results for closed-source ones (Appendix~E).

\begin{table}[ht!]
\centering
\caption{Comparison with generative data augmentation methods on Waterbirds and CelebA. The Group Info column indicates whether a method uses prior group labels to guide generation or a group-balanced validation set for model selection (\cmark), or operates without either (\xmark). Within each Group Info category, the best worst-group accuracy is underlined.}
\label{tab:generative_comparison}
\begin{tabular}{lccc}
\toprule
Method & Group Info & Waterbirds & CelebA \\
\cmidrule(lr){3-3} \cmidrule(lr){4-4}
& & WGA & WGA \\
\midrule
ASPIRE + ERM & \xmark & 78.7$_{\pm1.3}$ & 50.5$_{\pm0.8}$ \\
SAGE (ours) & \xmark & \underline{89.5$_{\pm0.1}$} & \underline{85.7$_{\pm0.3}$} \\
\midrule
ASPIRE + DFR & \cmark & 85.3$_{\pm1.3}$ & 85.5$_{\pm0.6}$ \\
DDB & \cmark & \underline{93.0$_{\pm0.1}$} & \underline{85.8$_{\pm1.4}$} \\
\makecell[l]{Clustered\\DreamBooth} & \cmark & 88.1$_{\pm0.9}$ & 84.1$_{\pm0.6}$ \\
\bottomrule
\end{tabular}
\end{table}

\begin{figure*}[t]
    \centering
    \resizebox{\textwidth}{!}{%
    \setlength{\tabcolsep}{3pt}%
    \renewcommand{\arraystretch}{1.1}%
    \begin{tabular}{c cccc cccc}
        & \multicolumn{4}{c}{\textbf{Waterbirds}} & \multicolumn{4}{c}{\textbf{CelebA}} \\
        \cmidrule(lr){2-5} \cmidrule(lr){6-9}
        & \makecell{\small landbird\\[-0.15em]\small $\langle t_5 \rangle$\\[-0.15em]\scriptsize (land bg)}
        & \makecell{\small landbird\\[-0.15em]\small $\langle t_8 \rangle$\\[-0.15em]\scriptsize (water bg)}
        & \makecell{\small waterbird\\[-0.15em]\small $\langle t_{150} \rangle$\\[-0.15em]\scriptsize (land bg)}
        & \makecell{\small waterbird\\[-0.15em]\small $\langle t_{177} \rangle$\\[-0.15em]\scriptsize (water bg)}
        & \makecell{\small not blond\\[-0.15em]\small $\langle t_5 \rangle$\\[-0.15em]\scriptsize (male)}
        & \makecell{\small not blond\\[-0.15em]\small $\langle t_{11} \rangle$\\[-0.15em]\scriptsize (female)}
        & \makecell{\small blond hair\\[-0.15em]\small $\langle t_{549} \rangle$\\[-0.15em]\scriptsize (female)}
        & \makecell{\small blond hair\\[-0.15em]\small $\langle t_{550} \rangle$\\[-0.15em]\scriptsize (male)} \\
        \makecell{\small ``a photo of [class label],\\[-0.15em]\small [sub-label token]''}
        & \raisebox{-0.5\height}{\includegraphics[width=1.55cm,height=1.55cm]{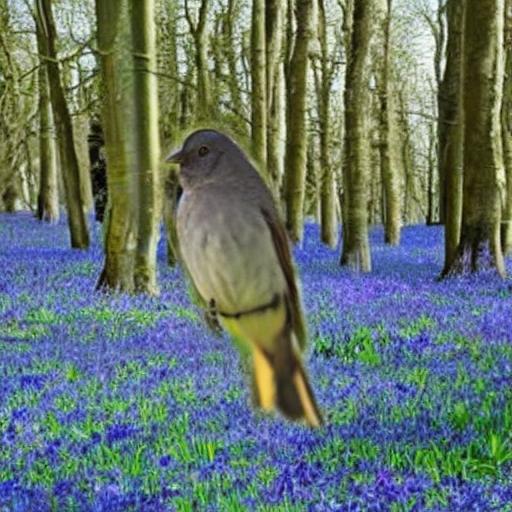}}
        & \raisebox{-0.5\height}{\includegraphics[width=1.55cm,height=1.55cm]{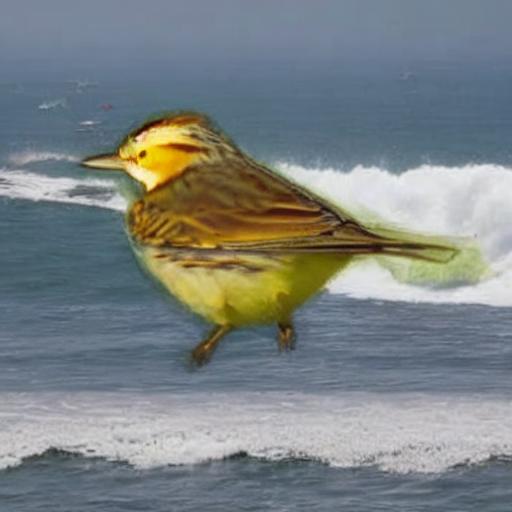}}
        & \raisebox{-0.5\height}{\includegraphics[width=1.55cm,height=1.55cm]{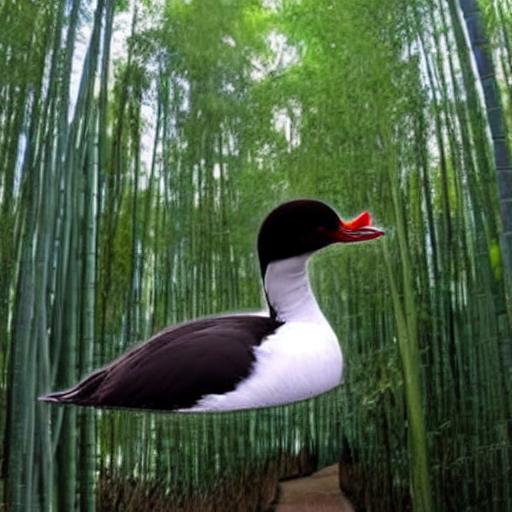}}
        & \raisebox{-0.5\height}{\includegraphics[width=1.55cm,height=1.55cm]{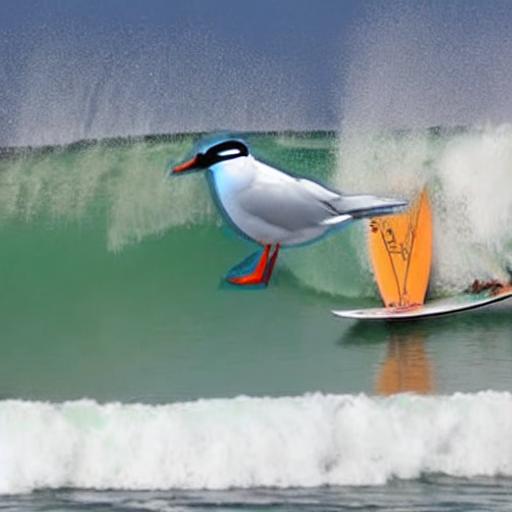}}
        & \raisebox{-0.5\height}{\includegraphics[width=1.55cm,height=1.55cm]{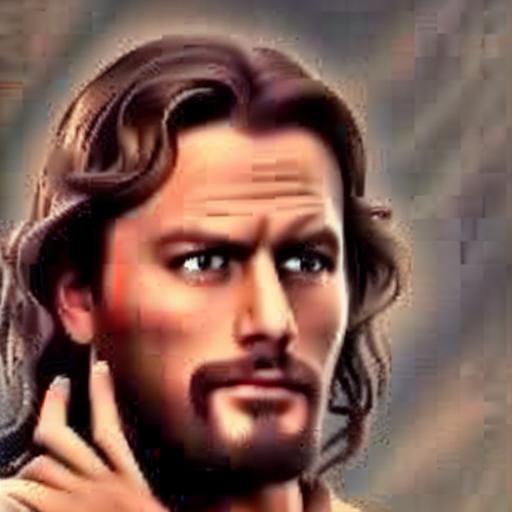}}
        & \raisebox{-0.5\height}{\includegraphics[width=1.55cm,height=1.55cm]{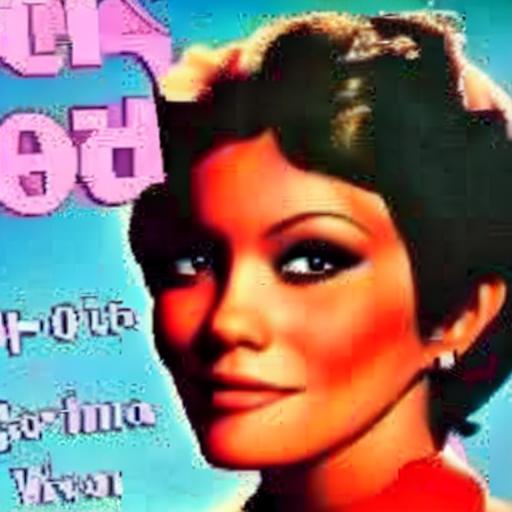}}
        & \raisebox{-0.5\height}{\includegraphics[width=1.55cm,height=1.55cm]{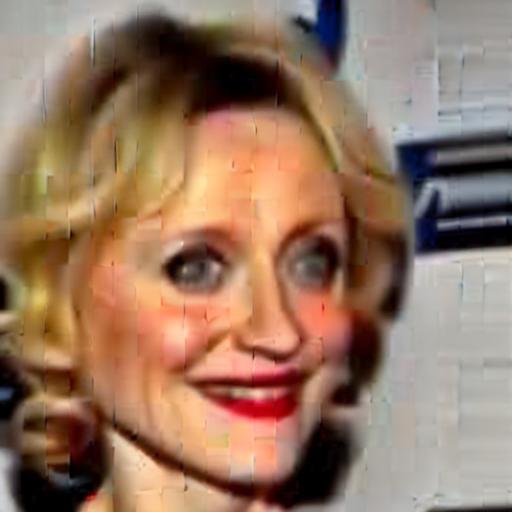}}
        & \raisebox{-0.5\height}{\includegraphics[width=1.55cm,height=1.55cm]{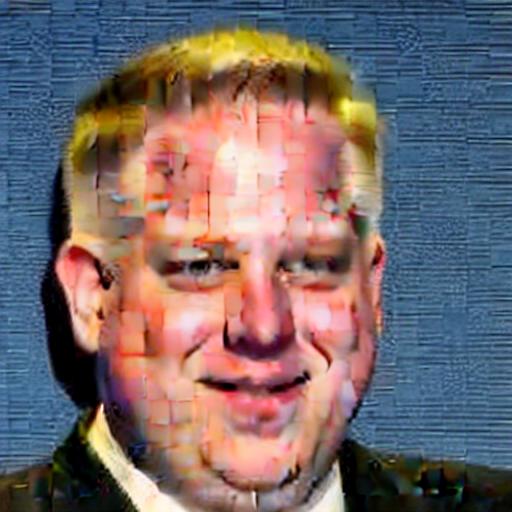}} \\[2.6em]
        \makecell{\small ``a photo of\\[-0.15em]\small [sub-label token]''}
        & \raisebox{-0.5\height}{\includegraphics[width=1.55cm,height=1.55cm]{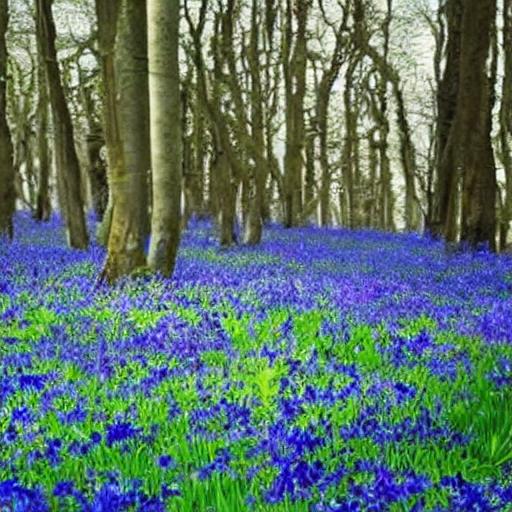}}
        & \raisebox{-0.5\height}{\includegraphics[width=1.55cm,height=1.55cm]{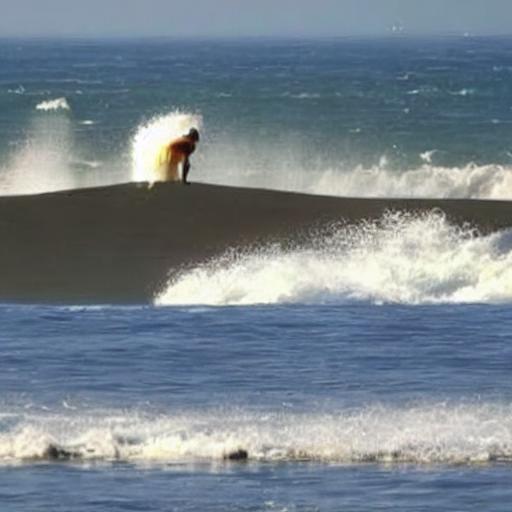}}
        & \raisebox{-0.5\height}{\includegraphics[width=1.55cm,height=1.55cm]{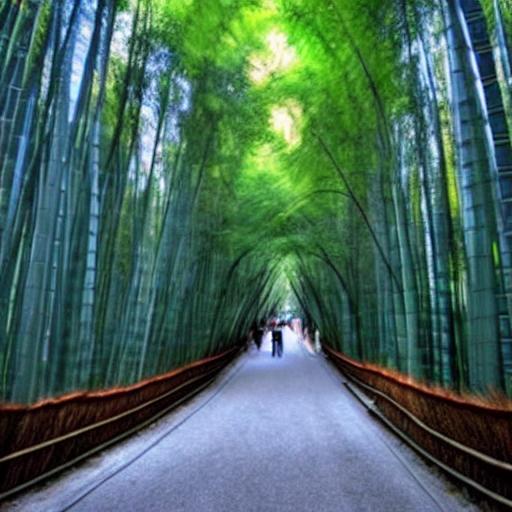}}
        & \raisebox{-0.5\height}{\includegraphics[width=1.55cm,height=1.55cm]{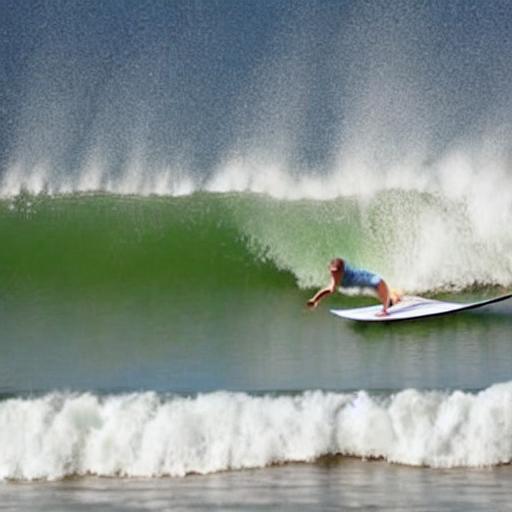}}
        & \raisebox{-0.5\height}{\includegraphics[width=1.55cm,height=1.55cm]{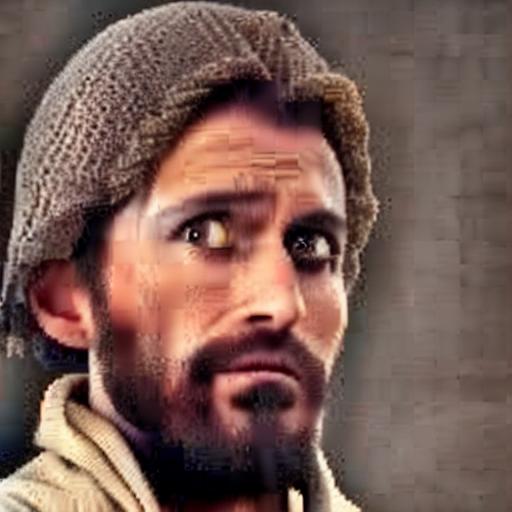}}
        & \raisebox{-0.5\height}{\includegraphics[width=1.55cm,height=1.55cm]{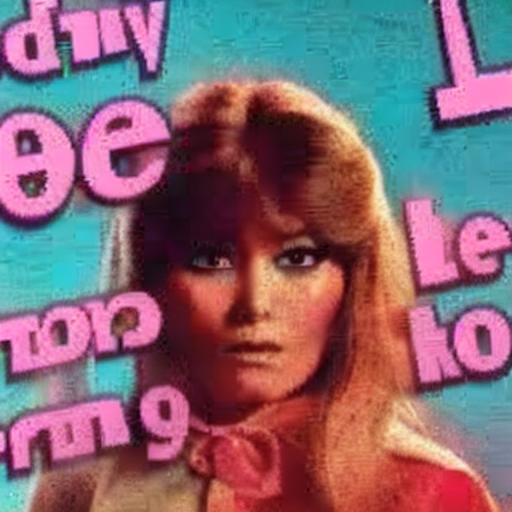}}
        & \raisebox{-0.5\height}{\includegraphics[width=1.55cm,height=1.55cm]{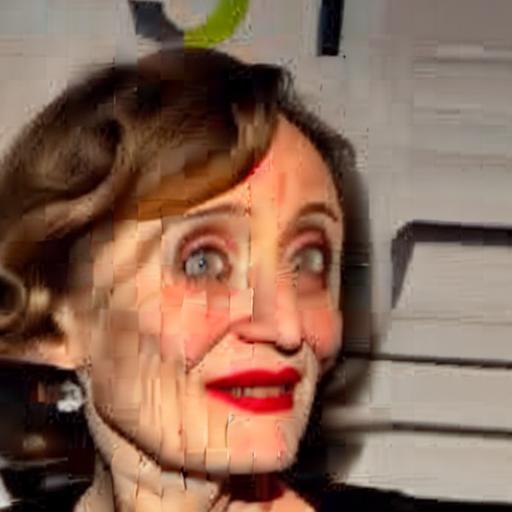}}
        & \raisebox{-0.5\height}{\includegraphics[width=1.55cm,height=1.55cm]{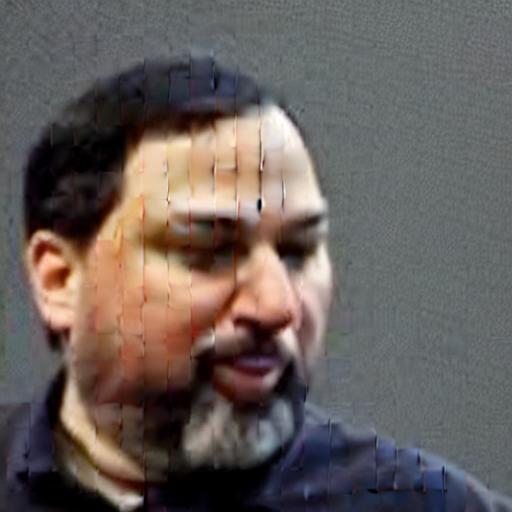}} \\
    \end{tabular}%
    }
    \caption{Qualitative validation of learned sub-label tokens on Waterbirds and CelebA, with column headers showing class label, token $\langle t_k \rangle$, and qualitatively exhibited visual factor. Top row: full prompt; fixing the class label while changing $\langle t_k \rangle$ alters background (Waterbirds) or gender (CelebA). Bottom row: omitting the class label still reproduces those visual factors from $\langle t_k \rangle$ alone (dataset-wide post-hoc cluster purity is quantified in Appendix~C).}
    \label{fig:token_visualization}
\end{figure*}

\subsection{Results}
\subsubsection{Compared with non-generative methods}
We report the performance of all compared baselines, the base ERM, and our proposed SAGE across three datasets. Baseline results for Waterbirds and CelebA are taken from the respective original papers, MetaShift results from \cite{noohdani2024decompose}, and JTT/LfF results without validation sets from \cite{asgari2022masktune}.

Table~\ref{tab:main_results} shows that, under the \emph{No group labels} setting (no group labels in training or validation), SAGE achieves the highest worst-group accuracy on all three benchmarks, exceeding the best baseline in this category by 0.8, 7.7, and 5.6 percentage points on Waterbirds, CelebA, and MetaShift, respectively. On CelebA, balancing heavily majority-dominated groups reflects a classic robustness--utility trade-off (average accuracy 95.5\% for ERM vs.\ 90.1\% for SAGE alongside +41.1 points WGA), whereas Waterbirds and MetaShift improve on both metrics. Notably, MetaShift tests only on unseen ``shelf'' backgrounds and has the lowest cluster purity (72.5\%, Appendix~C), yet SAGE attains the best WGA and average accuracy overall (79.1\% / 82.3\% vs.\ 64.2\% / 77.4\% for ERM), demonstrating tolerance to imperfect clustering under domain shift.

\subsubsection{Compared with generative methods}
For the generative data augmentation comparison, we report results on Waterbirds and CelebA, the two widely used benchmarks where all compared methods report results (FFR \cite{qraitem2024ffr} is excluded due to its CelebA-HQ smiling vs.\ gender protocol and attribute-named prompts). Table~\ref{tab:generative_comparison} compares SAGE with existing generative data augmentation methods. SAGE outperforms ASPIRE+ERM---the only other generative approach without manual group annotations---on both datasets, and surpasses ASPIRE+DFR without a manually curated validation set. Although SAGE slightly trails DDB, which relies on external segmentation, its simplicity and independence from external models make it a strong alternative.

\subsubsection{Extended Evaluations (Appendix~E)}
Appendix~E reports extended evaluations: (i) ISIC medical imaging with subtle co-occurring artifacts (hair, gel bubbles), where SAGE achieves 57.8$\pm$3.3\% AUROC vs.\ 55.1$\pm$2.3\% (DISC), 38.9$\pm$1.5\% (ERM+aug), and 36.4$\pm$0.9\% (GroupDRO); (ii) visual encoder purity ablations (CLIP vs.\ ResNet-50 and DINOv2, Appendix~C); (iii) RTX~4090 GPU-hour accounting ($\mathcal{D}_{\mathrm{syn}}$ adds $\sim$6\%/16\%/6\% generator cost, Appendix~B); and (iv) matched-size DFR with balanced real training images vs.\ generated $\mathcal{D}_{\mathrm{val}}$ (80.8/69.0/67.9\% vs.\ 85.9/79.5/73.4\%).

\subsection{Ablation Studies}
We conduct ablation studies to assess individual components of SAGE, including conditional generation under different sub-label tokens, the mixed training set in Stage 2, and the synthetic reweighting set for DFR.

\subsubsection{Visual Insights from Generated Images}
In this subsection, we analyze conditional generation under different sub-label tokens to examine the subpopulation-specific visual factors captured by the learned tokens.

We use Waterbirds and CelebA for qualitative visualization. Figure~\ref{fig:token_visualization} shows that changing the sub-label token alters generated images while the class label is fixed in the full prompt, and $\langle t_k \rangle$ alone still reproduces the associated factor when the class label is omitted. On Waterbirds, the illustrated class-token combinations correspond post hoc to landbird/land, landbird/water, waterbird/land, and waterbird/water. Notably, $\langle t_8 \rangle$ and $\langle t_{150} \rangle$ correspond to visually conflicted clusters with only 25 training samples each; such rare combinations are upweighted by inverse-density sampling, consistent with reducing background reliance in the downstream classifier.

\subsubsection{Effect of the Mixed Training Set}
To isolate the performance gains brought solely by our data augmentation, we conduct an experiment excluding the final DFR step in Stage 2. We further disentangle our adapted generator from generic generative augmentation by introducing a control setting where vanilla Stable Diffusion v1.5 generates synthetic data using the standard prompt ``a photo of [class label]''. These data are combined with $\mathcal{D}$ at the same total size as $\mathcal{D}_{\mathrm{mix}}$, forming $\mathcal{D}_{\mathrm{sg}}$ (Table~\ref{tab:ablation_mixed_set}), using a fixed random seed and \cite{sagawa2020distributionally} hyperparameters for Waterbirds and CelebA.
\begin{table}[ht!]
    \centering
    \caption{Ablation on mixed training set without DFR. Each entry is worst-group accuracy (\%) of a ResNet-50 ERM classifier trained on the column dataset. $\mathcal{D}$ is the original training set; $\mathcal{D}_{\mathrm{sg}}$ adds generic Stable Diffusion v1.5 augmentation at the same total scale as $\mathcal{D}_{\mathrm{mix}}$ (see text above); $\mathcal{D}_{\mathrm{mix}} = \mathcal{D} \cup \mathcal{D}_{\mathrm{syn}}$ uses targeted SAGE synthesis. Parentheses give the WGA gain of $\mathcal{D}_{\mathrm{mix}}$ over $\mathcal{D}$.}
    \label{tab:ablation_mixed_set}
    \begin{tabular}{lccc}
        \toprule
        Dataset & $\mathcal{D}$ & $\mathcal{D}_{\mathrm{sg}}$ & $\mathcal{D}_{\mathrm{mix}}$ \\
        \midrule
        Waterbirds & 73.2 & 70.5 & 77.5 (+4.3) \\
        CelebA     & 44.6 & 44.1 & 49.8 (+5.2) \\
        MetaShift  & 64.2 & 65.6 & 70.3 (+6.1) \\
        \bottomrule
    \end{tabular}
\end{table}

As shown in Table~\ref{tab:ablation_mixed_set}, generic synthetic data ($\mathcal{D}_{\mathrm{sg}}$) offers limited benefit or harms WGA, whereas $\mathcal{D}_{\mathrm{mix}}$ yields consistent gains (+4.3 to +6.1 points), confirming that improvements stem from targeted subpopulation augmentation rather than generative augmentation alone. Because $\mathcal{D}_{\mathrm{sg}}$ alters both fine-tuning and prompt conditioning, Appendix~E further isolates sampling strategies (75.6/76.1/77.5\% WGA) and synthetic volumes $Q$ (65.9/73.2/77.5\%) under the fixed fine-tuned generator on Waterbirds, estimates minority-share increases (3.8\%/1.2\% $\rightarrow$ 21.6\%/12.5\%), and isolates $\mathcal{D}_{\mathrm{mix}}$ feature gains via fixed official-validation DFR (+0.6$\pm$1.0 / +18.5$\pm$1.4 / +7.9$\pm$0.4 pp).

\begin{table}[ht!]
    \centering
    \caption{Evaluation of the synthetic reweighting set $\mathcal{D}_{\mathrm{val}}$ on ERM backbones trained solely on $\mathcal{D}$ (3-seed mean WGA, \%). Both Orig and Gen use the same single-run DFR variant omitting repeated subsampling and $\ell_1$ regularization.}
    \label{tab:valset_results}
    \begin{tabular}{l c c c c}
        \toprule
        \multicolumn{5}{c}{\textbf{Waterbirds}} \\
        \midrule
        ERM WGA (\%) & \multicolumn{4}{c}{73.2} \\
        \midrule
        & Orig $\mathcal{D}_{\mathrm{val}}$ & \multicolumn{3}{c}{Gen $\mathcal{D}_{\mathrm{val}}$} \\
        \cmidrule(lr){2-2} \cmidrule(lr){3-5}
        Samples-per-token & - & 8 & 12 & 16 \\
        Reweighting Size  & 532  & 312  & 468  & 624  \\
        DFR WGA (\%) & 89.7 & 84.1 & 88.6 & 85.9 \\
        
        \midrule
        \multicolumn{5}{c}{\textbf{CelebA}} \\
        \midrule
        ERM WGA (\%) & \multicolumn{4}{c}{44.6} \\
        \midrule
        & Orig $\mathcal{D}_{\mathrm{val}}$ & \multicolumn{3}{c}{Gen $\mathcal{D}_{\mathrm{val}}$} \\
        \cmidrule(lr){2-2} \cmidrule(lr){3-5}
        Samples-per-token & - & 8 & 12 & 16 \\
        Reweighting Size  & 728  & 376  & 564  & 752  \\
        DFR WGA (\%) & 62.8 & 80.2 & 82.6 & 79.5 \\
        
        \midrule
        \multicolumn{5}{c}{\textbf{MetaShift}} \\
        \midrule
        ERM WGA (\%) & \multicolumn{4}{c}{64.2} \\
        \midrule
        & Orig $\mathcal{D}_{\mathrm{val}}$ & \multicolumn{3}{c}{Gen $\mathcal{D}_{\mathrm{val}}$} \\
        \cmidrule(lr){2-2} \cmidrule(lr){3-5}
        Samples-per-token & - & 2 & 4 & 6 \\
        Reweighting Size  & 68   & 44   & 88  & 132  \\
        DFR WGA (\%)  & 69.5 & 62.6 & 68.5 & 73.4 \\
        \bottomrule
    \end{tabular}
\end{table}

\subsubsection{Effect of the Synthetic Reweighting Set}
To evaluate our synthetic reweighting set $\mathcal{D}_{\mathrm{val}}$ independently of $\mathcal{D}_{\mathrm{mix}}$, Table~\ref{tab:valset_results} contrasts DFR on ResNet-50 backbones trained solely on $\mathcal{D}$ using either a manually balanced group split from the original validation set (Orig) or Gen $\mathcal{D}_{\mathrm{val}}$ at three `samples-per-token' scales (Appendix~B), under identical single-run DFR hyperparameters omitting repeated subsampling and $\ell_1$ regularization. Notably, Table~\ref{tab:main_results} uses `samples-per-token' $= 16 / 16 / 6$ fixed a priori without group labels or test feedback (yielding 85.9 / 79.5 / 73.4\% for ERM($\mathcal{D}$)+Gen $\mathcal{D}_{\mathrm{val}}$ vs.\ 89.5 / 85.7 / 79.1\% for full SAGE; Appendix~E), rather than the post-hoc Table~\ref{tab:valset_results} peaks (e.g., 12 gives 88.6 / 82.6\% on Waterbirds/CelebA).

On Waterbirds, where spurious attributes align cleanly with annotated $(y, a)$ groups, Orig remains strongest (89.7\% vs.\ up to 88.6\% Gen). On CelebA and MetaShift, which exhibit more heterogeneous, naturally occurring shortcuts, Gen $\mathcal{D}_{\mathrm{val}}$ reaches 82.6\% vs.\ 62.8\% (CelebA) and 73.4\% vs.\ 69.5\% (MetaShift). We attribute this to finer within-group sub-label balance (Appendix~C) and to majority-group downsampling variance in single-run Orig \cite{kirichenko2023last}---confirmed against balanced real training-image resampling in Appendix~E. Finally, Gen $\mathcal{D}_{\mathrm{val}}$ is sensitive to `samples-per-token': MetaShift at 2 samples per token obtains 62.6\%, below ERM (64.2\%), before rising to 68.5\% (4) and 73.4\% (6).

\section{Conclusion}
We propose SAGE, a two-stage method designed to mitigate spurious correlations without group labels, external discovery tools, or manually curated validation sets. Driven by a data-centric philosophy, SAGE uses cluster-derived sub-label tokens to adapt a single conditional generator, simultaneously expanding underrepresented regions ($\mathcal{D}_{\mathrm{mix}}$) and synthesizing a balanced synthetic reweighting set ($\mathcal{D}_{\mathrm{val}}$) for last-layer DFR. Experimental results across Waterbirds, CelebA, MetaShift, and ISIC medical imaging (Appendix~E) demonstrate strong efficacy and generalization potential across subpopulation-shift datasets.

\bibliography{aaai2027}

\clearpage
\FloatBarrier

This is the supplementary material of the paper ``SAGE: Subpopulation-Aware Generative Enhancement for Mitigating Spurious Correlations''. The content catalog is as follows:
\begin{itemize}
    \item Appendix A: Dataset Statistics
    \begin{itemize}
        \item Appendix A.1: Dataset Groups and Label Combinations
        \item Appendix A.2: Group Distributions by Split
    \end{itemize}
    \item Appendix B: Implementation Details and Computational Cost
    \begin{itemize}
        \item Appendix B.1: Denoising Conditional Diffusion Model Details
        \item Appendix B.2: Hyperparameters
        \item Appendix B.3: Hardware and Computational Overhead
    \end{itemize}
    \item Appendix C: Clustering Analysis
    \begin{itemize}
        \item Appendix C.1: Clustering Hyperparameters
        \item Appendix C.2: Cluster Statistics and Visual Encoder Ablation
        \item Appendix C.3: Clustering Strategy for CelebA
    \end{itemize}
    \item Appendix D: Effectiveness Analysis of DFR with the Synthetic Reweighting Set
    \begin{itemize}
        \item Appendix D.1: Feature-Space Construction
        \item Appendix D.2: Downstream Classification Analysis
    \end{itemize}
    \item Appendix E: Extended Experiments and Analyses
    \begin{itemize}
        \item Appendix E.1: Evaluation on Medical Imaging (ISIC)
        \item Appendix E.2: Isolating the Contribution of Mixed Training ($\mathcal{D}_{\mathrm{mix}}$)
        \item Appendix E.3: Synthetic Generation vs.\ Real-Image Resampling for DFR
        \item Appendix E.4: Controlled Generator, Sampling, Volume, and Minority-Coverage Analyses on Waterbirds
        \item Appendix E.5: Baseline Provenance, FFR Exclusion, and Limitations
    \end{itemize}
\end{itemize}

\section{Appendix A: Dataset Statistics}
\label{app:datasets}

We report group definitions and split-wise $(y, a)$ distributions for the three benchmarks used in the main paper. All counts are computed from the dataset metadata in our experiments. SAGE performs clustering and generative fine-tuning on the training split only (4,795 / 162,770 / 1,024 images on Waterbirds, CelebA, and MetaShift, respectively); benchmark validation splits are not used in the method pipeline, and test samples are held out for final evaluation. Appendix~C reports post-hoc clustering purity on the union of training and validation splits (5,994 / 182,637 / 1,105 images) for numerical analysis only. Implementation details and clustering analysis are given in Appendix~B and Appendix~C, respectively.

\subsection{Dataset Groups and Label Combinations}
\label{app:datasets:groups}

Figure~\ref{fig:dataset_overview} summarizes the four $(y, a)$ groups per benchmark. Each column $\mathcal{G}_i$ shows a square-cropped training example, a short group description, the class label $y$, and the number of training and validation images. Example images are center-cropped and resized to a common square size. The figure is placed at the end of this supplementary material.

\FloatBarrier
\subsection{Group Distributions by Split}
\label{app:datasets:distribution}

Tables~\ref{tab:wb_dist}--\ref{tab:ms_dist} report the number of images in each $(y, a)$ group on the training, validation, and test splits.

\begin{table}[!ht]
\centering
\small
\begin{tabular}{lccc}
\toprule
Group $(y, a)$ & Train & Val & Test \\
\midrule
landbird, land background   & 3,498 & 467   & 2,255 \\
landbird, water background  & 184   & 466   & 2,255 \\
waterbird, land background  & 56    & 133   & 642   \\
waterbird, water background & 1,057 & 133   & 642   \\
\bottomrule
\end{tabular}
\caption{Waterbirds group distribution by split ($N_{\mathrm{train}}\!=\!4{,}795$, $N_{\mathrm{val}}\!=\!1{,}199$, $N_{\mathrm{test}}\!=\!5{,}794$). Validation and test splits are approximately group-balanced.}
\label{tab:wb_dist}
\end{table}

\begin{table}[!ht]
\centering
\small
\begin{tabular}{lccc}
\toprule
Group $(y, a)$ & Train & Val & Test \\
\midrule
not blond, female & 71,629 & 8,535 & 9,767 \\
not blond, male   & 66,874 & 8,276 & 7,535 \\
blond, female     & 22,880 & 2,874 & 2,480 \\
blond, male       & 1,387  & 182   & 180   \\
\bottomrule
\end{tabular}
\caption{CelebA group distribution by split ($N_{\mathrm{train}}\!=\!162{,}770$, $N_{\mathrm{val}}\!=\!19{,}867$, $N_{\mathrm{test}}\!=\!19{,}962$).}
\label{tab:celebA_dist}
\end{table}

\begin{table}[!ht]
\centering
\small
\begin{tabular}{lccc}
\toprule
Group $(y, a)$ & Train & Val & Test \\
\midrule
cat, bed    & 348 & 0   & 0   \\
cat, sofa   & 202 & 0   & 0   \\
dog, bench  & 124 & 0   & 0   \\
dog, bike   & 350 & 0   & 0   \\
cat, shelf  & 0   & 34  & 201 \\
dog, shelf  & 0   & 47  & 259 \\
\bottomrule
\end{tabular}
\caption{MetaShift group distribution by split ($N_{\mathrm{train}}\!=\!1{,}024$, $N_{\mathrm{val}}\!=\!81$, $N_{\mathrm{test}}\!=\!460$). Shelf background appears only in validation and test.}
\label{tab:ms_dist}
\end{table}

\FloatBarrier
\section{Appendix B: Implementation Details and Computational Cost}
\label{app:implementation}

This appendix supplements the Method and Experiment sections of the main paper with reproducibility-oriented details and computational overhead accounting. Dataset split sizes and group distributions are given in Appendix~A; clustering-specific hyperparameters and post-hoc purity analysis are reported separately in Appendix~C.

\subsection{B.1 Denoising Conditional Diffusion Model Details}
\label{app:impl-diffusion}

SAGE builds on Hugging Face Stable Diffusion v1.5\footnote{\url{https://huggingface.co/stable-diffusion-v1-5/stable-diffusion-v1-5}}, which performs iterative denoising in the latent space of a frozen Variational Autoencoder (VAE) rather than directly in pixel space:
\begin{itemize}
    \item \textbf{Latent representation and training:} Each training image $x$ is encoded into $z = \mathcal{E}(x)$ by the frozen VAE encoder. During training under \texttt{DDPMScheduler}, Gaussian noise $\epsilon \sim \mathcal{N}(0, \mathbf{I})$ is added at timestep $t \sim \mathcal{U}(1, T)$ to form the noisy latent $z_t$.
    \item \textbf{Denoising U-Net and cross-attention conditioning:} The U-Net takes $(z_t, t)$ and prompt embeddings $\tau(y, e_k)$ as input and predicts the added noise $\hat{\epsilon}$. In cross-attention layers, spatial image features provide the queries ($Q$), while encoded prompt embeddings provide the keys ($K$) and values ($V$). Prompts follow the template \textit{``a photo of [class label $y$], [sub-label token $\langle t_k \rangle$]''}, so conditioning specifies both class $y$ and discovered subpopulation $k$.
    \item \textbf{Inference and decoding:} At generation time, starting from Gaussian noise $z_T \sim \mathcal{N}(0, \mathbf{I})$, the adapted U-Net iteratively predicts and removes noise over 50 denoising steps using the pipeline's default \texttt{PNDMScheduler} and classifier-free guidance scale $7.5$, and the frozen VAE decoder maps the denoised latent back to a $512 \times 512$ image.
\end{itemize}

\subsection{B.2 Hyperparameters}
\label{app:impl-hyperparams}

Unless stated otherwise, we use the same configuration across Waterbirds, CelebA, and MetaShift.

\paragraph{Conditional generator adaptation (Stable Diffusion v1.5 + LoRA).}
We adapt the U-Net with Low-Rank Adaptation (LoRA) on attention projections (\texttt{to\_q}, \texttt{to\_k}, \texttt{to\_v}, \texttt{to\_out.0}) and jointly optimize newly registered sub-label token embeddings $\langle t_k \rangle$ initialized from the word \textit{photo}. A gradient mask ensures only the newly added token embeddings receive updates in the text encoder, while all remaining text-encoder, VAE, and base U-Net weights stay frozen. Key settings are:
\begin{itemize}
    \item LoRA rank $r = 128$; LoRA $\alpha = 128$; mixed precision \texttt{fp16}
    \item Optimizer: AdamW; U-Net LoRA lr: $10^{-5}$; token lr: $10^{-3}$
    \item Batch size (per GPU): 4; epochs: 100; resolution: $512 \times 512$
    \item Sampling during diffusion training: token-balanced weighted sampling across clusters
    \item Generation: 50 denoising steps; guidance scale $7.5$
\end{itemize}

\paragraph{Token similarity merging.}
After fine-tuning, we merge redundant sub-label tokens within each class (main paper, Stage~2). For class $y$, let $\mathbf{e}_k$ denote the learned text-encoder embedding of sub-label token $\langle t_k \rangle$. We compute all pairwise cosine similarities $\cos(\mathbf{e}_i, \mathbf{e}_j)$ among tokens in that class and take their mean $\bar{s}_y$. The merge criterion is not a fixed absolute threshold; instead, two tokens are merged if
\begin{equation}
    \cos(\mathbf{e}_i, \mathbf{e}_j) > \tau_{\mathrm{merge}} = \beta \cdot \bar{s}_y,
\end{equation}
where $\beta$ is a dataset-specific multiplier (shared across classes within the same dataset). Merging is applied transitively to form disjoint semantic groups. We set $\beta = 2.5$ on Waterbirds, $4.0$ on CelebA, and $4.0$ on MetaShift, fixed a priori without using group labels or test feedback.

\paragraph{Synthetic reweighting-set synthesis ($\mathcal{D}_{\mathrm{val}}$).}
We empirically observe that when synthesizing a small-scale reweighting set, relying on purely stochastic probability sampling can introduce substantial variance in the resulting data distribution. Consequently, for synthetic reweighting-set construction, we replace stochastic inverse-density sampling with a deterministic \emph{samples-per-token} count per merged group. In Table~\ref{tab:main_results} of the main paper, \emph{samples-per-token} is fixed a priori to $16$, $16$, and $6$ on Waterbirds, CelebA, and MetaShift, respectively, before running the Table~\ref{tab:valset_results} sensitivity study and without using group labels or test feedback.

\subsection{B.3 Hardware and Computational Overhead}
\label{app:Hardware}

Stable Diffusion LoRA adaptation is run on two NVIDIA RTX~4090 GPUs (24\,GB VRAM capacity each). All other stages---CLIP feature extraction, clustering, synthetic image generation, ResNet-50 training, and DFR---are run on a single NVIDIA RTX~4090 GPU.

Because Stage~1 LoRA adaptation is already required to generate the synthetic reweighting set $\mathcal{D}_{\mathrm{val}}$, Table~\ref{tab:compute_cost} summarizes the measured GPU-hours of LoRA adaptation ($t_{\mathrm{LoRA}}$) and generating $\mathcal{D}_{\mathrm{syn}}$ for mixed training ($t_{\mathrm{syn}}$). Relative to their combined generator cost $t_{\mathrm{LoRA}} + t_{\mathrm{syn}}$, generating $\mathcal{D}_{\mathrm{syn}}$ adds only $\sim$6\%, $\sim$16\%, and $\sim$6\% overhead on Waterbirds, CelebA, and MetaShift, respectively.

\begin{table}[!ht]
\centering
\small
\begin{tabular}{lccc}
\toprule
Dataset & LoRA $t_{\mathrm{LoRA}}$ & $\mathcal{D}_{\mathrm{syn}}$ $t_{\mathrm{syn}}$ & Added Share \\
\midrule
Waterbirds & $\sim$16 GPU-h  & $\sim$1 GPU-h   & $\sim$6\%  \\
CelebA     & $\sim$128 GPU-h & $\sim$24 GPU-h  & $\sim$16\% \\
MetaShift  & $\sim$8 GPU-h   & $\sim$0.5 GPU-h & $\sim$6\%  \\
\bottomrule
\end{tabular}
\caption{Generator computational cost (GPU-hours on 24\,GB RTX~4090 GPUs). Additional cost share is $t_{\mathrm{syn}} / (t_{\mathrm{LoRA}} + t_{\mathrm{syn}})$.}
\label{tab:compute_cost}
\end{table}

\section{Appendix C: Clustering Analysis}
\label{app:clustering}

\subsection{C.1 Clustering Hyperparameters}

We apply Affinity Propagation (AP) clustering independently within each class using L2-normalized image embeddings extracted from a frozen CLIP ViT-B/32 image encoder \cite{radford2021learning}. Pairwise similarity is defined as the negative squared Euclidean distance $s(i,k) = -\|z_i - z_k\|^2$. The AP algorithm has two key hyperparameters: the damping factor $\lambda_{\mathrm{damp}}$, which controls numerical oscillation during message passing, and the diagonal preference values $s(k, k)$, set to the mean of all off-diagonal pairwise similarities scaled by $\alpha_{\mathrm{pref}}$. For Waterbirds and MetaShift, whose training sets are small enough for exact per-class AP, we set $\lambda_{\mathrm{damp}} = 0.5$. For CelebA, which contains over 160,000 training samples, we increase damping to $\lambda_{\mathrm{damp}} = 0.8$ to ensure stable convergence. Across all three datasets, we fix $\alpha_{\mathrm{pref}} = 1.0$.

\subsection{C.2 Cluster Statistics and Visual Encoder Ablation}

Ground-truth spurious labels are used only in this post-hoc analysis and are not available to SAGE during training or generation. For purity evaluation, we re-cluster on the union of training and validation splits (Table~\ref{tab:cluster_counts}); this is separate from the train-split-only clustering used in the method. For datasets with such labels, we evaluate \textit{clustering purity}---the extent to which discovered clusters align with spurious attributes. Let $\mathcal{C}=\{C_1,\dots,C_K\}$ denote the set of discovered clusters and $\mathcal{A}$ the set of spurious attributes ($|\mathcal{A}| = 2$ for Waterbirds and CelebA; $|\mathcal{A}| = 4$ training attributes on MetaShift, while the 81 validation images have the unseen \emph{shelf} background as noted in Table~\ref{tab:ms_dist}). For each cluster $C_k$, its purity is defined as:
\begin{equation}
    \mathrm{Purity}(C_k) = \max_{a} \frac{|\{(x_i, y_i, a_i) \in C_k : a_i = a\}|}{|C_k|}.
\end{equation}
The overall purity reported in Table~\ref{tab:cluster_counts} is the sample-weighted (micro) average over clusters:
\begin{equation}
    \mathrm{Purity}_{\mathrm{overall}} = \frac{1}{\sum_{k=1}^{K}|C_k|}\sum_{k=1}^{K}|C_k|\,\mathrm{Purity}(C_k).
\end{equation}

\begin{table}[!ht]
\centering
\begin{tabular}{lccc}
\toprule
Dataset & Samples & Clusters & Purity (\%) \\
\midrule
Waterbirds & 5,994 & 180 & 93.66 \\
CelebA     & 182,637 & 1,089 & 98.17 \\
MetaShift  & 1,105 & 113 & 72.49 \\
\bottomrule
\end{tabular}
\caption{Number of samples, resulting clusters, and clustering purity for each dataset (CLIP ViT-B/32). Sample counts equal the sum of training and validation splits and are used for post-hoc analysis only.}
\label{tab:cluster_counts}
\end{table}

As shown in Table~\ref{tab:cluster_counts}, purity exceeds 90\% on Waterbirds and CelebA, indicating that clusters often align with binary spurious attributes on these datasets. On MetaShift, purity is lower (72.49\%) because four training attributes (sofa, bed, bench, bike) plus the unseen validation background (shelf) present finer variation and because cats and dogs exhibit strong inter-class visual cues competing with background features. Despite this weaker alignment, SAGE still achieves the best WGA and average accuracy on MetaShift (Table~\ref{tab:main_results}), showing tolerance to imperfect clustering.

\paragraph{Visual encoder ablation.}
To assess whether Stage~1 clustering depends specifically on CLIP, Table~\ref{tab:encoder_purity} compares cluster purity when replacing CLIP ViT-B/32 \cite{radford2021learning} with ResNet-50 \cite{he2016resnet} and DINOv2 \cite{oquab2024dinov2} image features. All three general-purpose visual encoders achieve high purity, with CLIP ViT-B/32 performing most consistently across the three benchmarks. (Note that Table~\ref{tab:encoder_purity} evaluates Stage~1 cluster purity; full downstream WGA runs for the alternative clustering backbones are left to future work.)

\begin{table}[!ht]
\centering
\small
\setlength{\tabcolsep}{3pt}
\begin{tabular}{@{}lccc@{}}
\toprule
Visual Encoder & Waterbirds & CelebA & MetaShift \\
\midrule
ResNet-50        & 93.1 & 87.4 & \textbf{72.6} \\
DINOv2        & 86.9 & 95.6 & 71.1 \\
CLIP ViT-B/32 & \textbf{93.7} & \textbf{98.2} & 72.5 \\
\bottomrule
\end{tabular}
\caption{Post-hoc cluster purity (\%) across visual encoders on Waterbirds, CelebA, and MetaShift.}
\label{tab:encoder_purity}
\end{table}

\subsection{C.3 Clustering Strategy for CelebA}

With over 160,000 training samples, CelebA poses a computational challenge for Affinity Propagation: storing the full $N \times N$ similarity matrix is infeasible. Whereas Waterbirds and MetaShift are small enough to run exact per-class AP directly, on CelebA we adopt a three-stage strategy.

First, we reduce the dataset to a diverse core set of $K = 10{,}000$ samples via Farthest Point Sampling (FPS) \cite{eldar1997farthest}, which iteratively selects points farthest from the already chosen set, ensuring broad coverage of the feature space. Second, we run AP clustering exclusively on this core set to identify exemplars. Third, we build a Faiss exact nearest-neighbor index \cite{johnson2021billion} over the exemplars and assign every sample in the full dataset to its closest exemplar, thereby recovering cluster labels for all samples.

We repeat this procedure separately for each class (blond / not blond) and offset cluster indices globally to prevent overlap. The dominant memory cost reduces from storing the $N \times N$ similarity matrix (infeasible at $N \approx 1.6\times 10^5$) to $O(N \times D + K^2)$, where $D$ is the CLIP feature dimension and $K \ll N$. In our experiments, $K$ is set to $10{,}000$.

\section{Appendix D: Effectiveness Analysis of DFR with the Synthetic Reweighting Set}
\label{app:decision-boundary}

\subsection{D.1 Feature-Space Construction}

\begin{figure*}[!t]
\centering
\setlength{\tabcolsep}{3pt}
\renewcommand{\arraystretch}{1.05}
\begin{tabular}{cc}
\textbf{Reweighting samples} & \textbf{Held-out test set} \\
\includegraphics[width=0.34\textwidth]{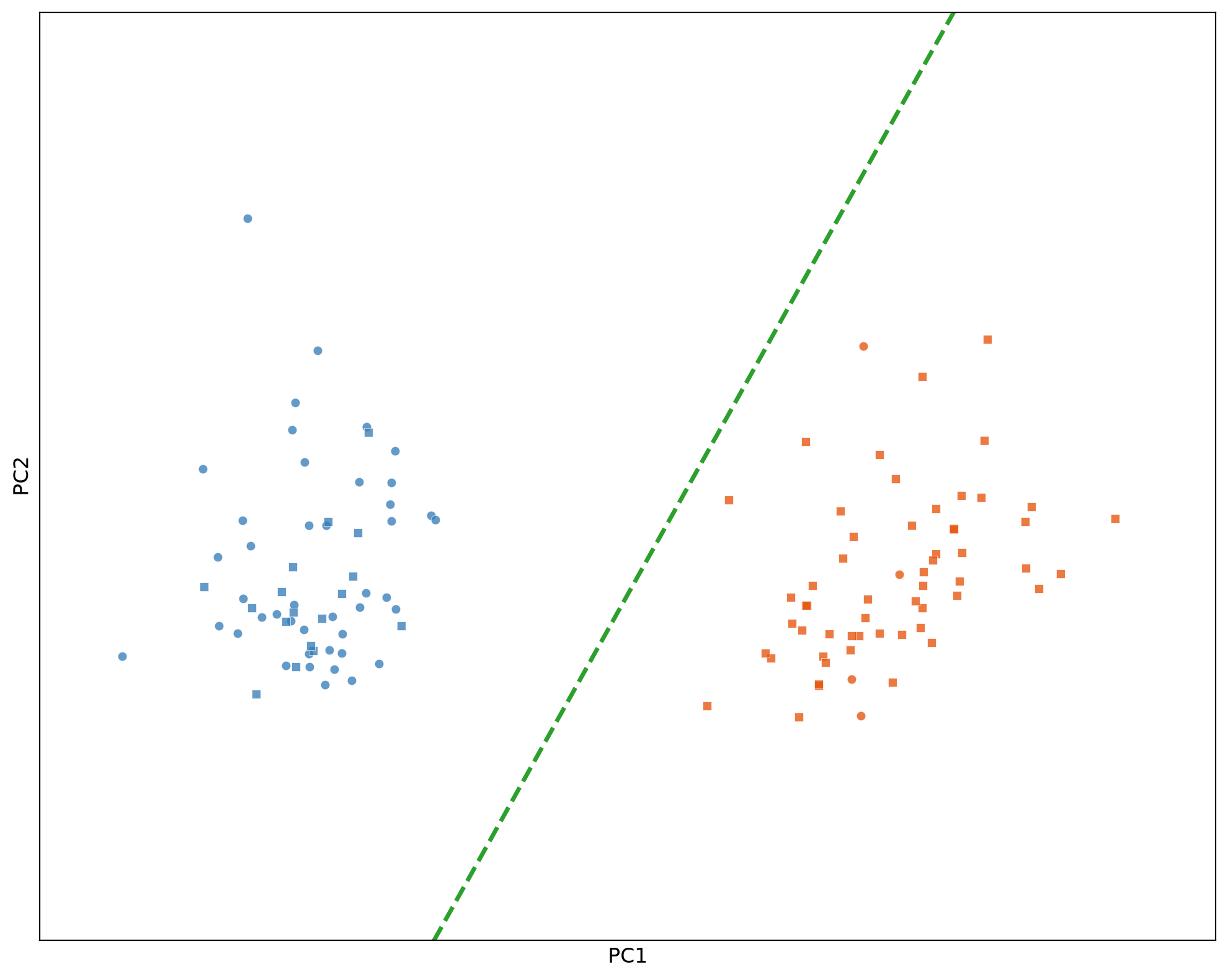} &
\includegraphics[width=0.34\textwidth]{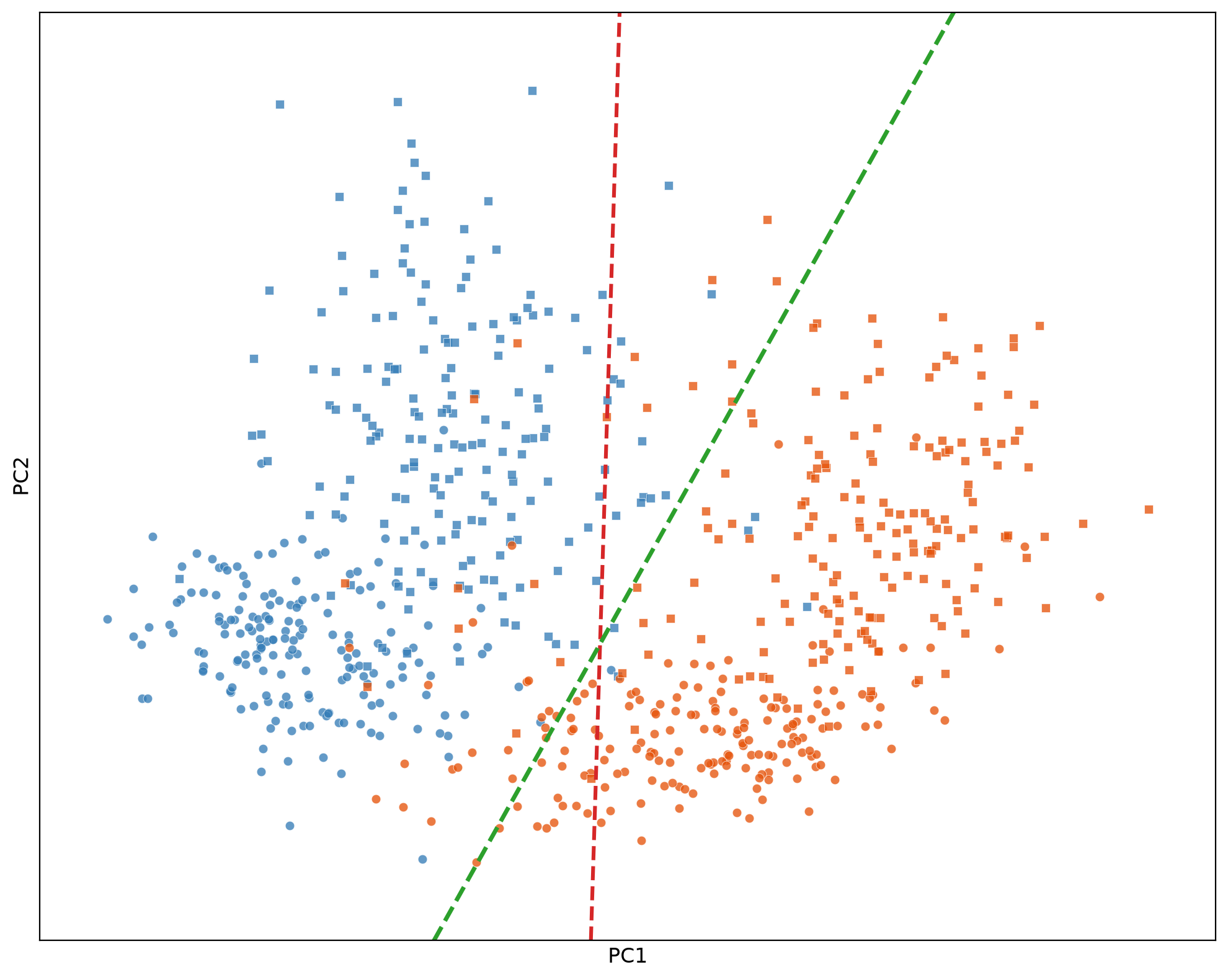} \\
{\scriptsize (a) Waterbirds reweighting samples} &
{\scriptsize (b) Waterbirds test set} \\[-0.2em]
\includegraphics[width=0.34\textwidth]{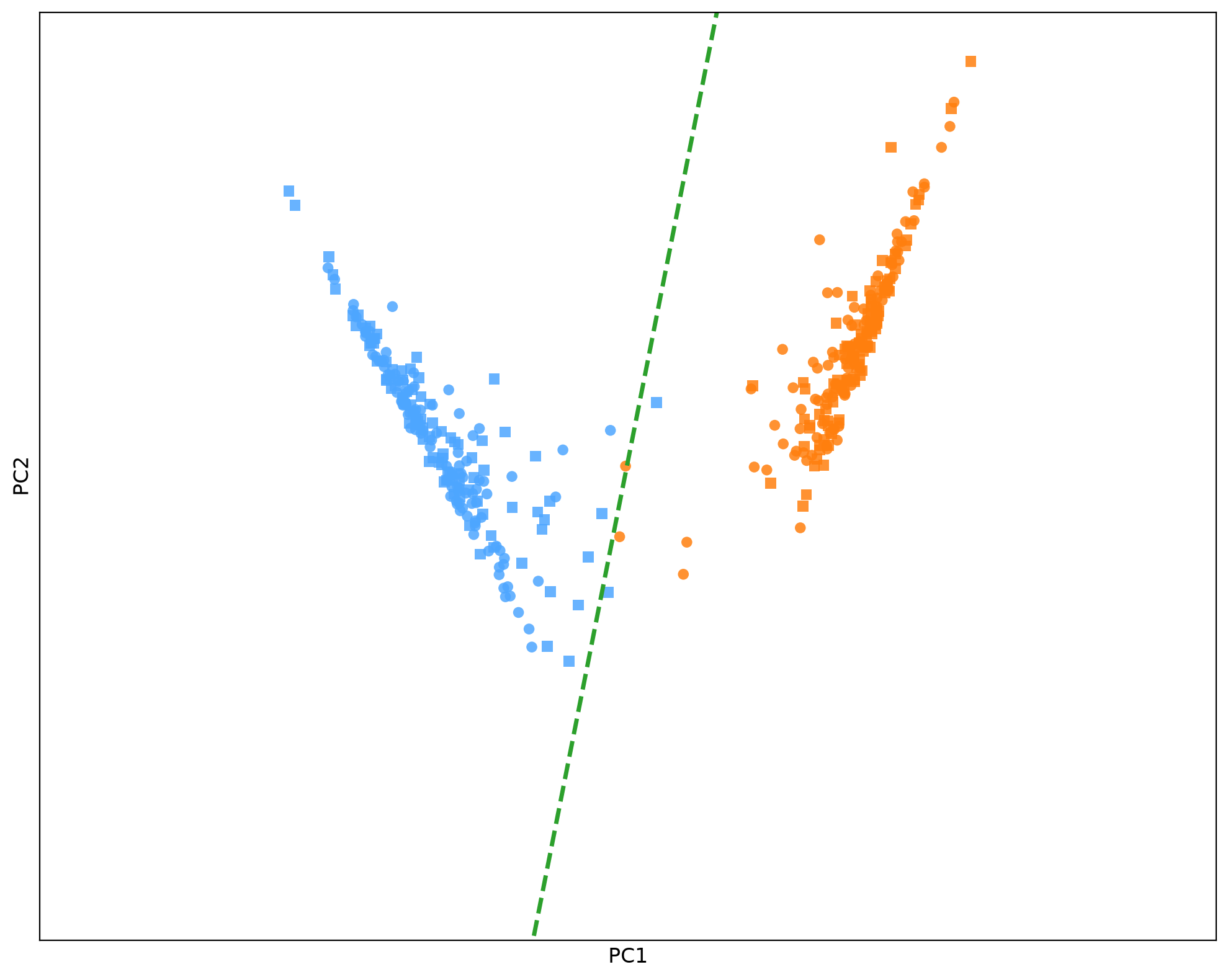} &
\includegraphics[width=0.34\textwidth]{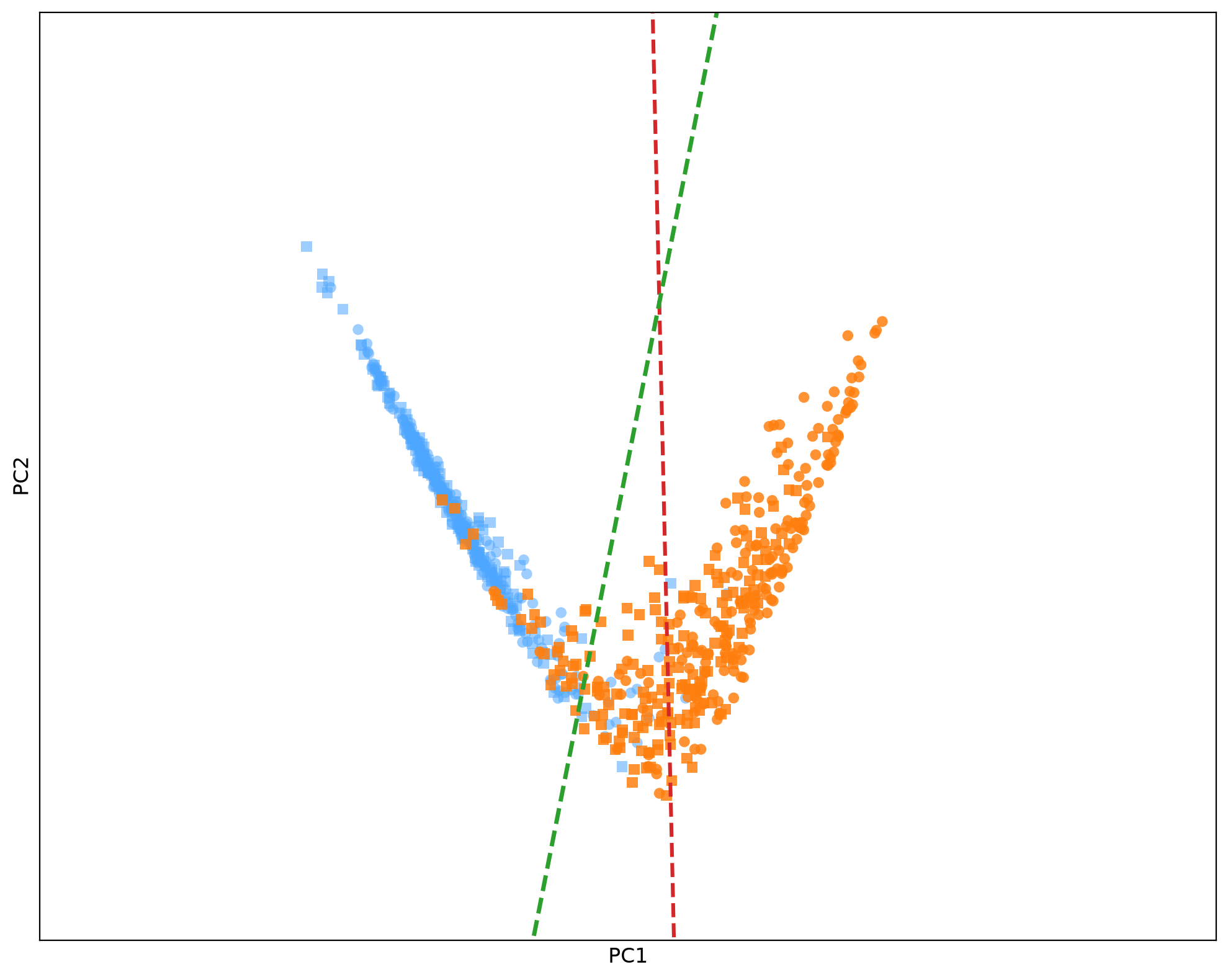} \\
{\scriptsize (c) CelebA reweighting samples} &
{\scriptsize (d) CelebA test set} \\[-0.2em]
\includegraphics[width=0.34\textwidth]{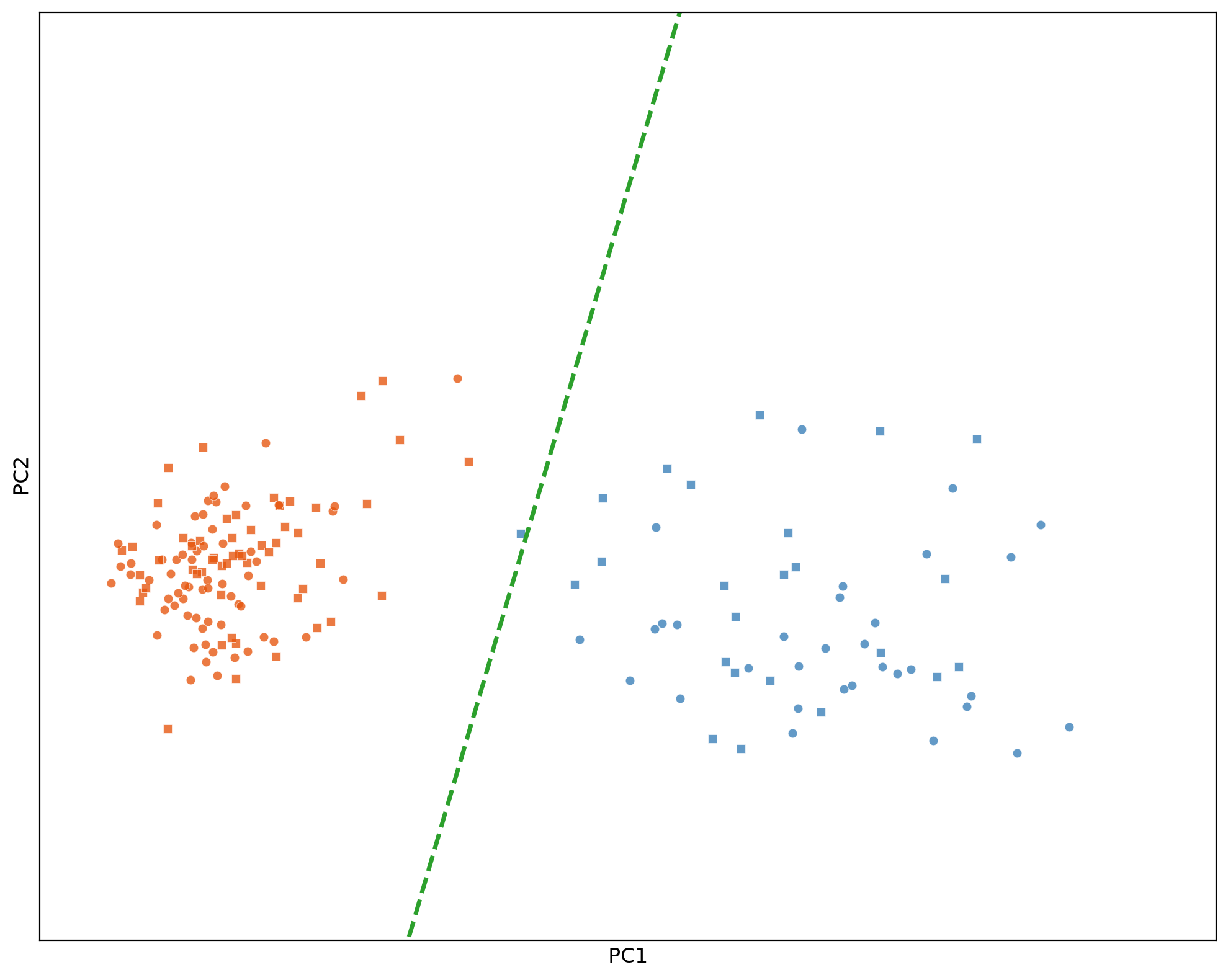} &
\includegraphics[width=0.34\textwidth]{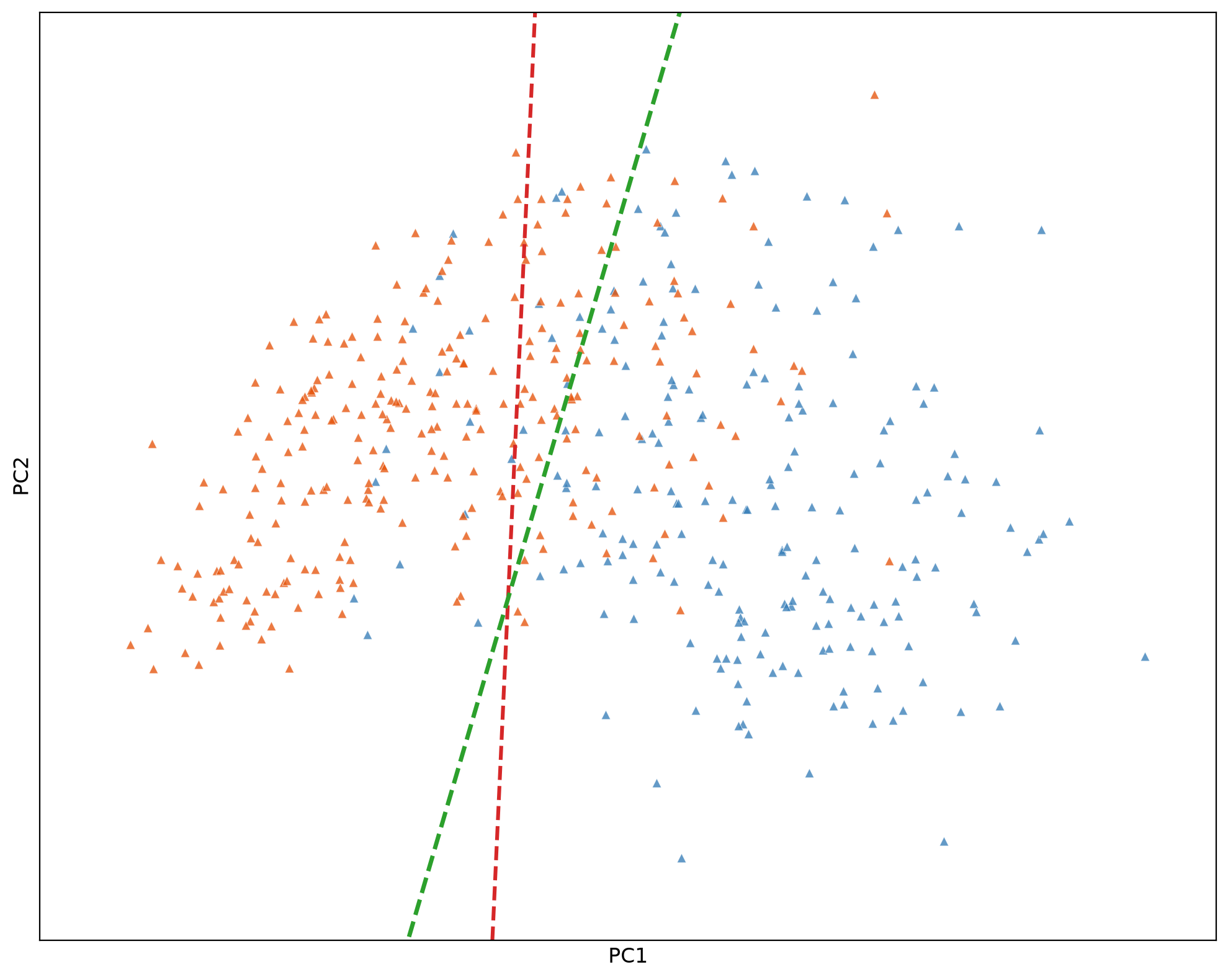} \\
{\scriptsize (e) MetaShift reweighting samples} &
{\scriptsize (f) MetaShift test set}
\end{tabular}
\caption{Decision-boundary visualizations for Waterbirds, CelebA, and MetaShift. Left column: synthetic reweighting samples from $\mathcal{D}_{\mathrm{val}}$ with the decision boundary after DFR. Right column: held-out test samples with both the ERM decision boundary and the decision boundary after DFR. Green dashed line: decision boundary after DFR; red dashed line: ERM decision boundary. Blue/orange points denote the two target classes: landbird/waterbird, non-blond/blond, and cat/dog, respectively. Circles/squares denote the available spurious attribute for Waterbirds and CelebA; for MetaShift, they denote merged generated-background groups in the left panel, while test samples are triangles because all come from the held-out shelf background.}
\label{fig:boundary_visualization}
\end{figure*}

Figure~\ref{fig:boundary_visualization} analyzes how applying DFR on the synthetic reweighting set constructed by SAGE affects downstream classification in the frozen feature space. For each dataset, we use the ERM-trained backbone as a fixed feature extractor. Let $h_\theta(\cdot)$ denote this backbone after replacing the final classification layer with an identity map, and let $f=h_\theta(x)\in\mathbb{R}^d$ be the frozen feature of an image $x$. We construct the two-dimensional PCA space separately for each dataset. Specifically, let $\mathcal{D}_{\mathrm{test}}^{\mathrm{vis}}$ denote the held-out test samples used in the visualization and let $\mathcal{D}_{\mathrm{val}}^{\mathrm{DFR}}$ denote the synthetic reweighting samples used for DFR. PCA is fit on the union of their frozen features,
\begin{equation}
\mathcal{S}_{\mathrm{PCA}}
= \{h_\theta(x): x\in\mathcal{D}_{\mathrm{test}}^{\mathrm{vis}}\}
\cup \{h_\theta(x): x\in\mathcal{D}_{\mathrm{val}}^{\mathrm{DFR}}\}.
\end{equation}
If $F=[f_1,\ldots,f_n]^\top$ stacks these features, we first center them by $\mu=\frac{1}{n}\sum_i f_i$ and compute the empirical covariance matrix
\begin{equation}
C=\frac{1}{n}\sum_{i=1}^{n}(f_i-\mu)(f_i-\mu)^\top .
\end{equation}
The PCA basis is given by the two eigenvectors with the largest eigenvalues, $u_1$ and $u_2$, satisfying $C u_j=\lambda_j u_j$ with $\lambda_1\geq\lambda_2$. For any frozen feature $f$, its two-dimensional coordinate is then
\begin{equation}
z = U^\top(f-\mu), \qquad U=[u_1,u_2]\in\mathbb{R}^{d\times 2}.
\end{equation}
We use the same mean $\mu$ and basis $U$ to project both the reweighting samples and the held-out test samples, so the two panels in each row share a common coordinate system. The PCA basis is fit only on $\mathcal{D}_{\mathrm{test}}^{\mathrm{vis}}$ and the synthetic reweighting subset $\mathcal{D}_{\mathrm{val}}^{\mathrm{DFR}}$ defined above; any additional generated reweighting samples shown in the left column are transformed afterward using this fixed PCA map. The left column contains the synthetic reweighting samples used for DFR, while the right column contains samples from the held-out test split. For Waterbirds and CelebA, whose original test splits are large, $\mathcal{D}_{\mathrm{test}}^{\mathrm{vis}}$ is a balanced subset with 180 images sampled from each $(y,a)$ group. This keeps the visualization readable while preserving the class-attribute structure needed for comparing decision boundaries. Finally, for a binary final linear layer with logits $w_c^\top f+b_c$, the decision boundary in the original feature space is
\begin{equation}
(w_1-w_0)^\top f + (b_1-b_0)=0 .
\end{equation}
Substituting the PCA-plane approximation $f\approx \mu+Uz$ gives the boundary drawn in Figure~\ref{fig:boundary_visualization},
\begin{equation}
\big((w_1-w_0)^\top U\big) z + (w_1-w_0)^\top\mu + (b_1-b_0)=0 .
\end{equation}
This allows us to compare the ERM decision boundary with the decision boundary after DFR while keeping the backbone representation fixed.

\subsection{D.2 Downstream Classification Analysis}

The six panels compare the proxy reweighting set and the held-out test distribution under the same frozen feature representation. Applying DFR with $\mathcal{D}_{\mathrm{val}}$ updates only the last linear layer, which shifts the decision boundary in the projected test feature space. This boundary shift is consistent with the role of $\mathcal{D}_{\mathrm{val}}$ as a more balanced proxy reweighting set: it supplies additional coverage for rare or under-represented class-attribute combinations and reduces the dependence of the final classifier on majority-dominated correlations. Together with the quantitative results reported in the main paper, this feature-space analysis supports the conclusion that DFR on the synthetic reweighting set improves worst-group accuracy while maintaining competitive average accuracy, i.e., the robustness gain is obtained without a large sacrifice in overall classification performance.

\FloatBarrier

\section{Appendix E: Extended Experiments and Analyses}
\label{app:extended}

This appendix presents the additional evaluations, component-isolation experiments, and protocol clarifications referenced in the main paper.

\subsection{E.1 Evaluation on Medical Imaging (ISIC)}
\label{app:isic}

To evaluate whether SAGE transfers beyond natural-image benchmarks where Stable Diffusion v1.5 has abundant pre-training exposure, we test on the ISIC skin-lesion dataset \cite{codella2019skin}, a medical imaging benchmark featuring subtle, co-occurring spurious artifacts such as hair and gel bubbles. Following the DISC data splits, ResNet-50 backbone, and AUROC evaluation protocol \cite{wu2023discover}, Table~\ref{tab:isic} compares SAGE against ERM+aug, GroupDRO (which uses domain labels), and DISC (best result reported in \cite{wu2023discover}). SAGE achieves 57.8$\pm$3.3\% AUROC without group or domain annotations, outperforming all compared baselines.

\begin{table}[!ht]
\centering
\small
\setlength{\tabcolsep}{3pt}
\begin{tabular}{@{}lcc@{}}
\toprule
Method & \makecell{Group or\\domain labels} & AUROC (\%) \\
\midrule
ERM + aug & \xmark & $38.9_{\pm1.5}$ \\
GroupDRO \cite{sagawa2020distributionally} & \cmark & $36.4_{\pm0.9}$ \\
DISC \cite{wu2023discover} & \xmark & $55.1_{\pm2.3}$ \\
\textbf{SAGE (ours)} & \xmark & $\mathbf{57.8_{\pm3.3}}$ \\
\bottomrule
\end{tabular}
\caption{Evaluation on ISIC medical imaging following the DISC splits, ResNet-50 backbone, and AUROC (\%) protocol \cite{wu2023discover}. Baseline numbers follow \cite{wu2023discover}.}
\label{tab:isic}
\end{table}

\subsection{E.2 Isolating the Contribution of Mixed Training ($\mathcal{D}_{\mathrm{mix}}$)}
\label{app:dmix_isolation}

Because synthetic DFR ($\mathcal{D}_{\mathrm{val}}$) accounts for a large share of SAGE's overall gain, we isolate the contribution of mixed training ($\mathcal{D}_{\mathrm{mix}}$) in two ways (Table~\ref{tab:dmix_isolation}). First, matching the Table~\ref{tab:main_results} \emph{samples-per-token} setting ($16 / 16 / 6$ on Waterbirds, CelebA, and MetaShift), we compare an ERM backbone trained only on $\mathcal{D}$ plus generated $\mathcal{D}_{\mathrm{val}}$ (from Table~\ref{tab:valset_results}) against full SAGE (ERM on $\mathcal{D}_{\mathrm{mix}}$ plus $\mathcal{D}_{\mathrm{val}}$): $\mathcal{D}_{\mathrm{mix}}$ improves final WGA by +3.6, +6.2, and +5.7 percentage points, respectively.

Second, to isolate the backbone representation quality learned from $\mathcal{D}_{\mathrm{mix}}$ independently of synthetic $\mathcal{D}_{\mathrm{val}}$, we fix the official benchmark validation set for DFR and swap only the frozen ERM backbone (trained on $\mathcal{D}_{\mathrm{mix}}$ vs.\ $\mathcal{D}$) over three seeds. Note that official validation group labels are used strictly for this post-hoc diagnostic. Replacing $\mathcal{D}$ with $\mathcal{D}_{\mathrm{mix}}$ changes DFR WGA by $+0.6\pm1.0$, $+18.5\pm1.4$, and $+7.9\pm0.4$ percentage points on Waterbirds, CelebA, and MetaShift, respectively---demonstrating substantial representation gains on CelebA and MetaShift and a modest, higher-variance improvement on Waterbirds.

\begin{table}[!ht]
\centering
\small
\setlength{\tabcolsep}{3pt}
\begin{tabular}{@{}lccc@{}}
\toprule
Variant & Waterbirds & CelebA & MetaShift \\
\midrule
ERM ($\mathcal{D}$) & 73.2 & 44.6 & 64.2 \\
ERM ($\mathcal{D}$) + Gen $\mathcal{D}_{\mathrm{val}}$ & 85.9 & 79.5 & 73.4 \\
\makecell[l]{\textbf{Full SAGE}\\($\mathcal{D}_{\mathrm{mix}}$ + Gen $\mathcal{D}_{\mathrm{val}}$)} & \textbf{89.5} & \textbf{85.7} & \textbf{79.1} \\
\midrule
$\Delta$ WGA (fixed val) & $+0.6_{\pm1.0}$ & $+18.5_{\pm1.4}$ & $+7.9_{\pm0.4}$ \\
\bottomrule
\end{tabular}
\caption{Isolating $\mathcal{D}_{\mathrm{mix}}$'s contribution (WGA, \%). Top three rows match the Table~\ref{tab:main_results} \emph{samples-per-token} setting ($16/16/6$). Bottom row reports the 3-seed WGA change ($\Delta$ over $\mathcal{D}$) when fixing the official validation set for DFR and swapping only the ERM backbone.}
\label{tab:dmix_isolation}
\end{table}

\subsection{E.3 Synthetic Generation vs.\ Real-Image Resampling for DFR}
\label{app:real_vs_gen}

Given Stage~1 pseudo-group labels, a natural non-generative alternative for DFR is balancing original training images across merged groups without diffusion synthesis. Using the exact same pseudo-group labels, merged groups, and per-group and total sample counts as Gen $\mathcal{D}_{\mathrm{val}}$ (at the Table~\ref{tab:main_results} \emph{samples-per-token} settings $16/16/6$), Table~\ref{tab:real_vs_gen} reports 3-seed mean WGA on ERM($\mathcal{D}$) backbones. Generated $\mathcal{D}_{\mathrm{val}}$ outperforms balanced real training images across all three benchmarks (85.9 / 79.5 / 73.4\% vs.\ 80.8 / 69.0 / 67.9\%). Under this matched-count protocol, balancing real images requires downsampling larger groups, which introduces selection variance among retained examples---an effect synthetic generation avoids.

\begin{table}[!ht]
\centering
\small
\setlength{\tabcolsep}{3pt}
\begin{tabular}{@{}lccc@{}}
\toprule
Reweighting Source & Waterbirds & CelebA & MetaShift \\
\midrule
Balanced real images & 80.8 & 69.0 & 67.9 \\
\textbf{Generated $\mathcal{D}_{\mathrm{val}}$ (SAGE)} & \textbf{85.9} & \textbf{79.5} & \textbf{73.4} \\
\bottomrule
\end{tabular}
\caption{3-seed mean WGA (\%) comparing DFR using balanced real training images vs.\ generated $\mathcal{D}_{\mathrm{val}}$ under identical pseudo-groups and matched sample counts.}
\label{tab:real_vs_gen}
\end{table}

\subsection{E.4 Controlled Generator, Sampling, Volume, and Minority-Coverage Analyses on Waterbirds}
\label{app:wb_ablations}

Table~\ref{tab:ablation_mixed_set} in the main paper compares $\mathcal{D}_{\mathrm{sg}}$ (vanilla Stable Diffusion v1.5 with class-only prompts) against $\mathcal{D}_{\mathrm{mix}}$, which simultaneously changes generator adaptation and prompt conditioning. To disentangle generator fine-tuning, label sampling strategy, and synthetic data volume $Q$ cleanly, we run two controlled ablations on Waterbirds following the Table~\ref{tab:ablation_mixed_set} ERM-only protocol (no DFR, retaining all real images $\mathcal{D}$).

\begin{table}[!ht]
\centering
\small
\setlength{\tabcolsep}{3pt}
\begin{tabular}{@{}lc@{\hspace{12pt}}lc@{}}
\toprule
Generator / sampling & WGA & Volume $Q$ & WGA \\
\cmidrule(r){1-2}\cmidrule(l){3-4}
Original $\mathcal{D}$ & 73.2 & $0$ & 73.2 \\
Vanilla SD v1.5 & 70.5 & $|\mathcal{D}|/4$ & 65.9 \\
FT + proportional & 75.6 & $|\mathcal{D}|/2$ & 73.2 \\
FT + uniform & 76.1 & $|\mathcal{D}|$ & \textbf{77.5} \\
\textbf{FT + inverse density} & \textbf{77.5} & & \\
\bottomrule
\end{tabular}
\caption{Controlled Waterbirds ablations following the ERM-only protocol from Table~\ref{tab:ablation_mixed_set} (WGA, \%). The left panel compares generators and sampling strategies at $Q=|\mathcal{D}|$ for augmented runs. FT denotes fine-tuned Stable Diffusion v1.5. The right panel varies synthetic volume $Q$ under the fine-tuned generator and inverse-density sampling, with $|\mathcal{D}|$ as the default.}
\label{tab:wb_controlled}
\end{table}

As shown in Table~\ref{tab:wb_controlled}, fixing the fine-tuned generator at $Q = |\mathcal{D}|$ reveals consistent gains over $\mathcal{D}$ and $\mathcal{D}_{\mathrm{sg}}$ across all sampling rules, with inverse-density sampling performing best (77.5\% vs.\ 76.1\% uniform and 75.6\% proportional). Varying $Q \in \{|\mathcal{D}|/4, |\mathcal{D}|/2, |\mathcal{D}|\}$ yields 65.9\%, 73.2\%, and 77.5\% WGA, supporting our default $Q = |\mathcal{D}|$. Furthermore, weighting generated counts by each source cluster's empirical minority proportion (a source-cluster estimate rather than per-image re-annotation of synthetic outputs), we estimate that minority-group shares in Waterbirds rise from 3.8\% to 21.6\% (landbirds on water) and 1.2\% to 12.5\% (waterbirds on land) in $\mathcal{D}_{\mathrm{mix}}$, explaining the Table~\ref{tab:ablation_mixed_set} WGA gain (73.2\% $\rightarrow$ 77.5\%).

\subsection{E.5 Baseline Provenance, FFR Exclusion, and Limitations}
\label{app:provenance}

\paragraph{Baseline provenance and protocol verification.}
All experiments follow standard dataset splits, an ImageNet-pretrained ResNet-50 backbone, and standard evaluation protocols. In Table~\ref{tab:main_results}, Waterbirds and CelebA baselines are cited from their original papers, MetaShift baselines from DaC \cite{noohdani2024decompose}, and group-label-free JTT/LfF results from MaskTune \cite{asgari2022masktune}; in Table~\ref{tab:isic}, ISIC baselines are cited from DISC \cite{wu2023discover}. For all baselines with publicly available code, we verified split, backbone, and metric consistency against our setup; closed-source baselines quote their published numbers.

\paragraph{Exclusion of FFR from Table~\ref{tab:generative_comparison}.}
Although FFR \cite{qraitem2024ffr} is closely related in pretraining on balanced synthetic images, it is omitted from Table~\ref{tab:generative_comparison} because both its benchmark protocol and supervision setting differ from SAGE: FFR reports neither Waterbirds nor standard CelebA (blond hair), evaluating instead on CelebA-HQ (smiling vs.\ gender) with varied bias ratios, and its balanced synthesis prompt (\textit{``a photo of \{bias\} \{class\}''}) requires explicit prior knowledge of the spurious attribute.

\paragraph{Limitations.}
We note three limitations of our current evaluation: (i) generated image quality is examined qualitatively via token-conditioning grids (Figure~\ref{fig:token_visualization}) and post-hoc cluster-to-attribute alignment (Appendix~C), without separate quantitative per-image class-fidelity or perceptual-diversity metrics (e.g., FID/LPIPS); (ii) the controlled sampling-strategy and $Q$-ratio ablations in Appendix~E.4 are reported on Waterbirds, and alternative visual encoders in Appendix~C report cluster purity rather than full downstream retraining; and (iii) Table~\ref{tab:compute_cost} reports GPU-hours for LoRA adaptation and $\mathcal{D}_{\mathrm{syn}}$ synthesis on 24\,GB RTX~4090s, leaving fine-grained stage-by-stage wall-clock and peak-VRAM profiling to future work.

\FloatBarrier

\newcommand{\datasetimg}[1]{\includegraphics[width=1.85cm,height=1.85cm]{#1}}

\begin{figure*}[!p]
\centering
\setlength{\tabcolsep}{4pt}
\renewcommand{\arraystretch}{1.15}

\textbf{Waterbirds dataset}\\[0.4em]
\begin{tabular}{l|cccc}
\hline
 & $\mathcal{G}_1$ & $\mathcal{G}_2$ & $\mathcal{G}_3$ & $\mathcal{G}_4$ \\
\hline
 &
 \datasetimg{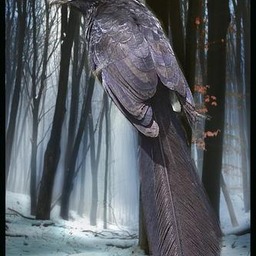} &
 \datasetimg{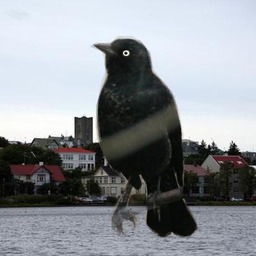} &
 \datasetimg{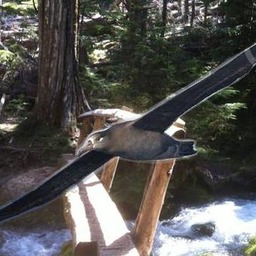} &
 \datasetimg{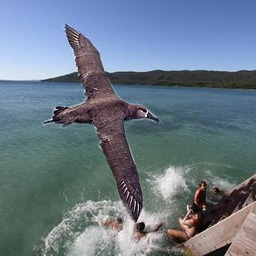} \\
Description &
 \makecell{landbird\\on land} &
 \makecell{landbird\\on water} &
 \makecell{waterbird\\on land} &
 \makecell{waterbird\\on water} \\
Class label & 0 & 0 & 1 & 1 \\
\# Train data & 3,498 & 184 & 56 & 1,057 \\
\# Val data & 467 & 466 & 133 & 133 \\
\hline
\end{tabular}\\[0.3em]
{\small\textit{Target:} bird type; \textit{Spurious feature:} background type; \textit{Minority:} $\mathcal{G}_2$, $\mathcal{G}_3$.}\\[1.2em]

\textbf{CelebA hair color dataset}\\[0.4em]
\begin{tabular}{l|cccc}
\hline
 & $\mathcal{G}_1$ & $\mathcal{G}_2$ & $\mathcal{G}_3$ & $\mathcal{G}_4$ \\
\hline
 &
 \datasetimg{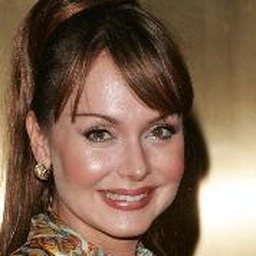} &
 \datasetimg{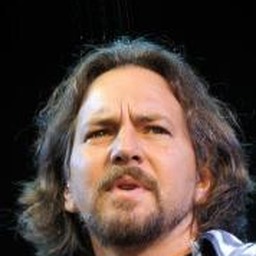} &
 \datasetimg{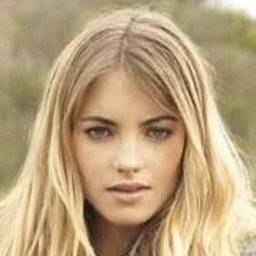} &
 \datasetimg{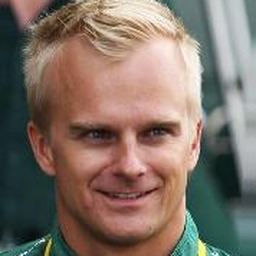} \\
Description &
 \makecell{non-blond\\woman} &
 \makecell{non-blond\\man} &
 \makecell{blond\\woman} &
 \makecell{blond\\man} \\
Class label & 0 & 0 & 1 & 1 \\
\# Train data & 71,629 & 66,874 & 22,880 & 1,387 \\
\# Val data & 8,535 & 8,276 & 2,874 & 182 \\
\hline
\end{tabular}\\[0.3em]
{\small\textit{Target:} hair color; \textit{Spurious feature:} gender; \textit{Minority:} $\mathcal{G}_4$.}\\[1.2em]

\textbf{MetaShift dataset}\\[0.4em]
\begin{tabular}{l|cccc}
\hline
 & $\mathcal{G}_1$ & $\mathcal{G}_2$ & $\mathcal{G}_3$ & $\mathcal{G}_4$ \\
\hline
 &
 \datasetimg{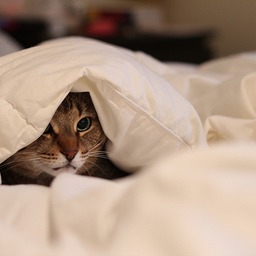} &
 \datasetimg{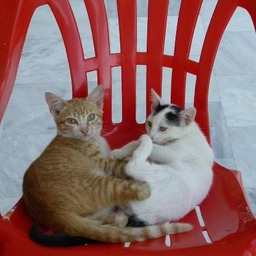} &
 \datasetimg{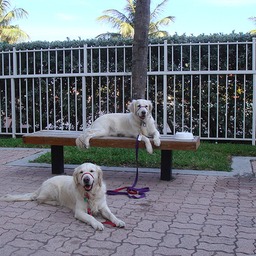} &
 \datasetimg{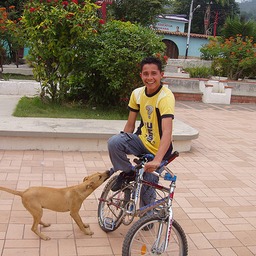} \\
Description &
 \makecell{cat\\on bed} &
 \makecell{cat\\on sofa} &
 \makecell{dog\\on bench} &
 \makecell{dog\\on bike} \\
Class label & 0 & 0 & 1 & 1 \\
\# Train data & 348 & 202 & 124 & 350 \\
\# Val data & 0 & 0 & 0 & 0 \\
\hline
\end{tabular}\\[0.3em]
{\small\textit{Target:} animal species; \textit{Spurious feature:} background object.}

\caption{Overview of the four $(y, a)$ groups for Waterbirds, CelebA, and MetaShift. Class label $y$ follows the standard group-ID convention in each benchmark.}
\label{fig:dataset_overview}
\end{figure*}

\FloatBarrier

\end{document}